\documentclass[10pt,twocolumn,letterpaper]{article}
\usepackage[pagenumbers]{cvpr} 

\usepackage{adjustbox}
\usepackage{multirow}
\usepackage[table]{xcolor}
\usepackage{bm}
\usepackage{booktabs}
\usepackage{makecell}
\usepackage{pifont}
\usepackage{lipsum}
\newcommand{\methodname}{ExposeAnyone\xspace}
\newcommand{\methodshortname}{EXAM\xspace}

\definecolor{cvprblue}{rgb}{0.21,0.49,0.74}
\usepackage[pagebackref,breaklinks,colorlinks,allcolors=cvprblue]{hyperref}

\title{\methodname: Personalized Audio-to-Expression Diffusion Models Are \\ Robust Zero-Shot Face Forgery Detectors}

\author{Kaede Shiohara$^{1}$  \quad Toshihiko Yamasaki$^{1}$ \quad Vladislav Golyanik$^{2}$  \qquad \vspace{1pt}\\
$^{1}$The University of Tokyo \qquad $^{2}$Max Planck Institute for Informatics, SIC\qquad\qquad\\
}

\begin{document}
\maketitle
\begin{abstract}

Detecting unknown deepfake manipulations remains one of the most challenging problems in face forgery detection.
Current state-of-the-art approaches fail to generalize to unseen manipulations, as they primarily rely on supervised training with existing deepfakes or pseudo-fakes, which leads to overfitting to specific forgery patterns.
In contrast, self-supervised methods offer greater potential for generalization, but existing work struggles to learn discriminative representations only from self-supervision.
In this paper, we propose \methodname, a fully self-supervised approach based on a diffusion model that generates expression sequences from audio.
The key idea is, once the model is personalized to specific subjects using reference sets,
it can compute the identity distances between suspected videos and personalized subjects via diffusion reconstruction errors, enabling person-of-interest face forgery detection.
Extensive experiments demonstrate that 1) our method outperforms the previous state-of-the-art method by 4.22 percentage points in the average AUC on DF-TIMIT, DFDCP, KoDF, and IDForge datasets, 
2) our model is also capable of detecting Sora2-generated videos, where the previous approaches perform poorly, and
3) our method is highly robust to corruptions such as blur and compression, highlighting the applicability in real-world face forgery detection\footnote{Project page: \url{https://mapooon.github.io/ExposeAnyonePage/}}.
\end{abstract}    
\begin{figure}[t]
  \centering
  \begin{adjustbox}{width=1.0\linewidth}
  \includegraphics{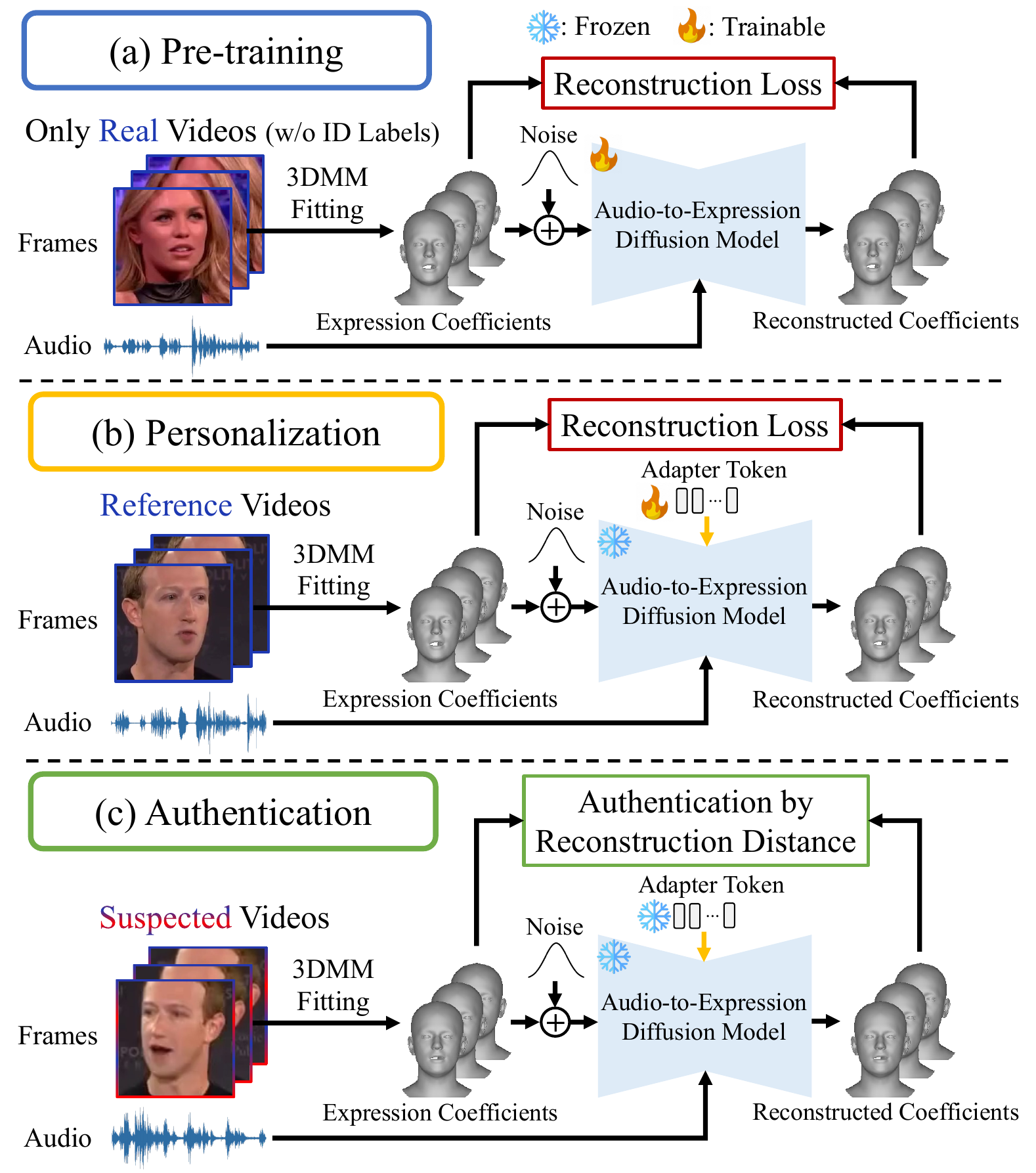}
  \end{adjustbox}
  \caption{\textbf{Our self-supervised face forgery detection approach:} We pre-train our audio-to-expression diffusion model on a large-scale, unlabeled video collection. Then, we personalize our pre-trained model on the reference videos of a person of interest (POI) by inserting a subject-specific adapter. Finally, we authenticate suspected videos of POI by the diffusion reconstruction distance.}
  \label{fig:teaser}
\end{figure}

\section{Introduction}
\label{sec:intro}
The recent advances in generative models such as generative adversarial networks~\cite{gan,dcgan,stylegan} and diffusion models~\cite{ddpm,ddim,ldm,imagen,kdiffusion} enable the generation of photo-realistic images and videos of human faces, which poses a risk of misuse for creating deceptive or malicious content, commonly referred to as \textbf{deepfakes}. 
It can undermine public trust, violate privacy, and spread misinformation; therefore, detection of deepfake media is an increasingly important research area. 
Recent studies have tackled this problem by reducing models' dependencies on manipulation-specific artifacts~\cite{ftcn,altfreezing,ucf,lipforensics}, synthesizing pseudo-fake samples that reproduce general artifacts observed in various deepfakes~\cite{fwa,facexray,pcl,sbi}, transferring large vision and language models~\cite{effort,forensicsadapter,dfdfcg}, and leveraging audio information for robust detection~\cite{realforensics,joint,liu2024lips}.
However, in general, it is well-known that these methods, which rely on learning forgery patterns from fake or pseudo-fake samples in training sets, inevitably become biased toward the seen manipulations and often fail to generalize to unseen ones~\cite{CanTheyBeGeneralized,cozzolino2018forensictransfer,xuan2019generalization,lae}.
To address the overfitting problem, some work stands on the perspective that detectors should be trained only on pristine data.
They quantify the plausibility of input videos based on reconstruction~\cite{ocfd}, audio-visual consistency~\cite{factor,avad,speechforensics}, and identity consistency with reference sets of specific subjects~\cite{idreveal,poiforensics}. 
However, they still struggle with learning discriminable representations only from self-supervision.
As a result, the performance of self-supervised approaches is very limited compared to state-of-the-art supervised and pseudo-supervised approaches.
In this paper, we propose \methodname, a new paradigm for face forgery detection that is independent of any actual fake or pseudo-fake samples yet detects face forgeries effectively. 
Our key idea is that diffusion models trained to generate person-specific facial expressions distinguish the real videos from fake videos that mimic the subject. 
We start by curating a large-scale audio-expression dataset, reaching 445 hours of data, with our proposed expression feature extraction strategy that disentangles face expression from face shape effectively.
Then, as shown in Fig.~\ref{fig:teaser}, we pre-train our simple diffusion model called \methodname Model (\methodshortname in short) based on Diffusion Transformer~\cite{dit} with several modifications for audio-to-expression generation.
After pre-training, we personalize \methodshortname to a subject by inserting a subject-specific adapter to learn the talking identity. 
Finally, we predict the fakeness of suspected videos, where the personalized subject appears, using our proposed content-agnostic authentication mechanism, which emphasizes the identity discrepancy.
Our method achieves 95.22\% in the average AUC on the traditional deepfake benchmarks~\cite{dftimit,dfdcp,kodf,idforge} and outperforms
the previous state-of-the-art approaches~\cite{efficientnet,facexray,lipforensics,ftcn,sbi,ict,realforensics,altfreezing,ucf,laanet,forensicsadapter,recce,liu2024lips,dfdfcg}.
Also, we introduce \textbf{Sora2 Cameo Forensics Preview}, the first dataset for Sora2-generated facial video detection. 
Our method is capable of detecting even Sora2-generated videos, while the previous methods perform poorly.
Moreover, we show that our method is highly robust to corruptions, especially a severe video compression rate, where the state-of-the-art AltFreezing method~\cite{altfreezing} drops performance by 36.71 percentage points in AUC, and ours drops just by 2.0 percentage points. 
Our contributions are summarized as follows:
\begin{enumerate}
\item We propose \methodname, a \textbf{fully self-supervised} face forgery detection framework based on \textbf{audio-to-expression diffusion models}. 
Our model is \textbf{personalized to specific subjects} to learn person-specific talking identity, which enables the model to expose fake videos with \textbf{state-of-the-art detection performance} on not only traditional deepfake videos but also Sora2-generated videos.
\item We introduce \textbf{a new 3DMM extraction strategy} and \textbf{content-agnostic authentication} in our framework. We validate in ablation studies that both are necessary for accurate detection. 
\item According to our extensive experiments, \methodname is the \textbf{only} self-supervised method that achieves competitive performance with the previous state-of-the-art approaches, suggesting a promising research direction for self-supervised face forgery detection.
\end{enumerate}

\section{Related Work}
\label{sec:relatedwork}
\noindent\textbf{Face Forgery Detection.}
Recent work tries to address the generalization problem revealed by early studies~\cite{CanTheyBeGeneralized,cozzolino2018forensictransfer,xuan2019generalization,lae}.
Some work tackles this problem by introducing general face boundary representation~\cite{facexray}, preventing overfitting to spatial information~\cite{ftcn,altfreezing}, leveraging pre-trained high-level semantics encoders~\cite{lipforensics,realforensics,forensicsadapter,effort,dfdfcg}, and combining audio cues~\cite{joint,avff,avlgi,liu2024lips}.
Other work focuses on training data synthesis, such as blended images~\cite{facexray,pcl,sladd} and self-blended images~\cite{sbi,ram,mtsbi} that reproduce artifacts generally seen in face swapping~\cite{deepfake-faceswap,faceswap,blendface,simswap,faceshifter} and face reenactment~\cite{neuraltextures,face2face}. 
Although they significantly improve baselines on existing benchmarks~\cite{ffpp,dfdcp,cdf}, the forgery types are limited within pre-defined patterns in training data, making it difficult to maintain detection performance on future manipulations.
Recent work focuses more on fully self-supervised methods that aim to achieve complete independence of overfitting to specific forgery patterns.
OC-FakeDect~\cite{ocfd} proposes reconstruction-based quantification of fakeness scores using variational auto-encoders~\cite{vae}.
SpeechForensics~\cite{avlipsyncplus} detects deepfake videos by computing the semantic similarity between voice signals and lip motions.
However, these methods still struggle to generalize due to the vast and unconstrained search space inherent to their self-supervised objectives, \eg, a single audio clip can correspond to various plausible visual patterns and vice versa, which makes it difficult to capture the audio-visual inconsistency on highly synchronized deepfake videos.

In contrast, we focus on the discrepancy of talking identity between real and fake subjects, which is manipulation-invariant and robust even on high-fidelity deepfake videos.

\begin{table}[t]
    \centering
    \begin{adjustbox}{width=1.0\linewidth}
    \begin{tabular}{lccc}
        \toprule
        Method & Self-Supervision & General Prior & Personalization\\ 
        \midrule
        PWL~\cite{agarwal2019protecting} & \checkmark& \ding{55} & \checkmark \\
        WTW~\cite{agarwal2023watch}  & \ding{55} & \ding{55} & \checkmark \\
        A\&B~\cite{agarwal2020detecting}       & \checkmark& \checkmark (w/ ID labels) & \ding{55}\\
        ID-Reveal~\cite{idreveal}& \checkmark& \checkmark (w/ ID labels) & \ding{55} \\
        POI-Forensics~\cite{poiforensics}    & \checkmark   & \checkmark (w/ ID labels) & \ding{55}\\
        STIDNet~\cite{fang2024stidnet}     & \checkmark  & \checkmark (w/ ID labels) & \ding{55}\\
         \rowcolor[gray]{0.9}
        Ours & \checkmark & \checkmark (\textbf{w/o} ID labels) & \checkmark\\
        \bottomrule
    \end{tabular}
    \end{adjustbox}
    \caption{\textbf{Concept-level comparison with previous reference-based methods.} Our method leverages general prior for personalization in a self-supervised fashion.}
    \label{tb:concept_comparison_reference}
\end{table}

\begin{figure*}[t]
  \centering
  \begin{adjustbox}{width=1.0\linewidth}
  \includegraphics{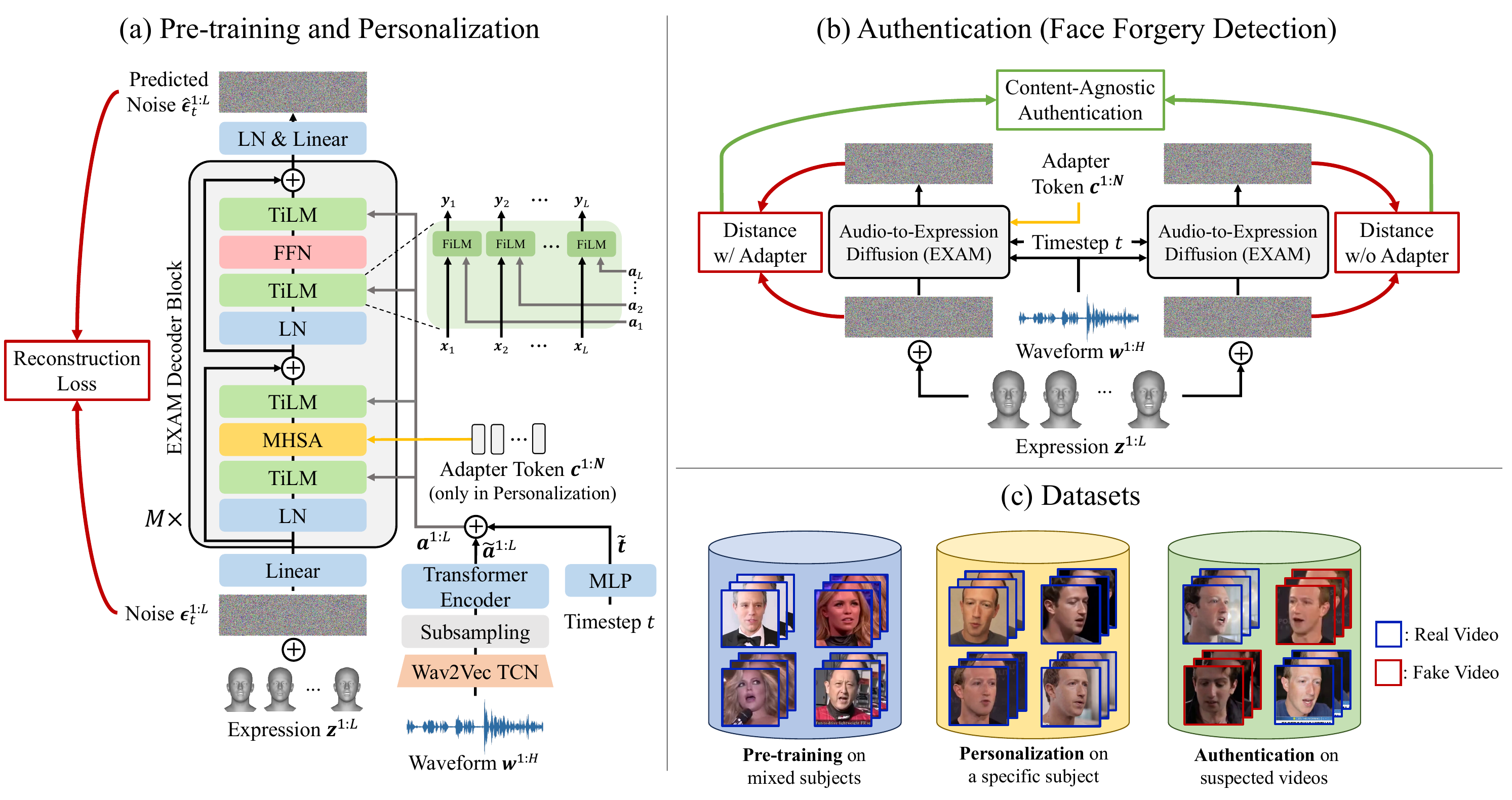}
  \end{adjustbox}
  \caption{\textbf{\methodname framework for face forgery detection.} (a) We pre-train an audio-to-expression diffusion model to predict the added noise sequence $\boldsymbol\epsilon^{1:L}$ from a noisy expression sequence ${\boldsymbol z}_t^{1:L}$. Then, we personalize the pre-trained model on a specific subject by inserting an adapter token sequence ${\boldsymbol c}^{1:K}$. (b) After personalization, our model can authenticate videos by computing two reconstruction distances w/ and w/o the adapter ${\boldsymbol c}^{1:K}$. (c) Our model is trained in a self-supervised fashion during both pre-training and personalization. 
  }
  \label{fig:method}
\end{figure*}

\noindent\textbf{Reference Assistance.}
Another research direction is reference-based approaches~\cite{idreveal,poiforensics,agarwal2019protecting,agarwal2023watch,agarwal2020detecting} that utilize reference videos, \ie, reliable pristine sources of subjects appearing in suspected videos.
We categorize previous work in Table~\ref{tb:concept_comparison_reference} with three important aspects: 
1) \textbf{Self-Supervision}, meaning training only on real videos.
2) \textbf{General Prior}, referring to the knowledge of generic facial dynamics learned from large-scale datasets consisting of non–person-of-interest individuals.
3) \textbf{Personalization}, meaning tuning on person-of-interest individuals.
From these aspects, 
WTW~\cite{agarwal2023watch} is trained with fake samples to classify whether input videos belong to a subject or not; therefore, it can be biased towards the seen manipulations.
PWL~\cite{agarwal2019protecting} and WTW~\cite{agarwal2023watch} are trained from scratch on each subject; therefore, they require a very long duration of reference videos due to the lack of general prior, \eg, PWL~\cite{agarwal2019protecting} requires 1 hour for each subject for accurate disentanglement. 
Other studies~\cite{idreveal,poiforensics,fang2024stidnet,agarwal2020detecting}
train models to extract identity features from videos and detect deepfakes by comparing identity similarity between suspected and reference videos, without personalization;
however, their discriminative ability is limited by the small number of identity-labeled subjects in existing video datasets,
\eg, the widely used VoxCeleb2~\cite{voxceleb2} provides only about 6K identities.

In contrast, our approach satisfies all three concepts as summarized in Table~\ref{tb:concept_comparison_reference}: 
1) It does not rely on any fake samples during training.
2) It learns a general prior from large-scale video collections without ID labels.
3) It is personalized to specific subjects, enabling the capture of more fine-grained identity-related facial dynamics.

\section{Proposed Framework: \methodname}
We give the overview of our framework in Fig.~\ref{fig:method}.
We first build a simple diffusion model, called \methodname Model (\methodshortname in short), that generates facial expressions from audio (Sec.~\ref{sec:model}). 
To train \methodshortname, we collect 200K videos with audio channels from publicly available datasets~\cite{voxceleb2,acappella,avspeech} and extract 3D facial expressions~\cite{flame} by introducing feed-forward initialization and iterative refinement (Sec.~\ref{sec:curation}). 
We pre-train the model on the curated dataset (Sec.~\ref{sec:pretraining}).
Then, we personalize our model on specific subjects to learn their talking identities (Sec.~\ref{sec:personalization}).
After that, our model can be used to authenticate suspected videos by comparing two reconstruction distances with and without identity information (Sec.~\ref{sec:authentication}).

\subsection{\methodname Model (\methodshortname)}
\label{sec:model}
We introduce a simple diffusion model, called \methodname Model (\methodshortname in short), that takes a speech signal and generates the corresponding facial expression sequence.

\noindent\textbf{Face Representation.} 
We adopt FLAME~\cite{flame}, a kind of 3D morphable models, that is widely incorporated in facial animation synthesis~\cite{voca,emote,gaussianavatars,headstudio}.
FLAME represents a face as a set of compressed coefficients $\boldsymbol\omega = (\boldsymbol\alpha, \boldsymbol\beta, \boldsymbol\gamma ) \in \mathbb{R}^{|\boldsymbol\alpha|+|\boldsymbol\beta|+|\boldsymbol\gamma|}$ for the face shape $\boldsymbol\alpha$,
expression $\boldsymbol\beta$,
and pose $\boldsymbol\gamma$.
\methodshortname aims to generate coefficients related to inner face motion, specifically, 50 principal expression parameters and three jaw pose parameters, denoted as $ {\boldsymbol z}\in \mathbb{R}^{53}$.

\noindent\textbf{Generator.}
We adopt denoising diffusion probabilistic models (DDPMs)~\cite{ddpm}, a class of state-of-the-art generative models~\cite{gan,stylegan,vae,vqvae,ldm}.
The overview of our model is shown in Fig.~\ref{fig:method}(a).
Because there is no public implementation of diffusion models to generate FLAME parameters solely from audio, we train a simple model based on Diffusion Transformer (DiT)~\cite{dit}, which is equipped mainly with Transformer~\cite{attention} encoder blocks consisting of multi-head self-attention (MHSA), layer normalization (LN)~\cite{layernorm}, and feed-forward network (FFN).
To encode audio information, we employ Wav2Vec 2.0~\cite{wav2vec}.
It is worth noting that we simply extend the feature-wise linear modulation (FiLM)~\cite{film} to sequential modeling, which we name time- and feature-wise linear modulation (TiLM).
Specifically, for an input sequence $\boldsymbol{x}^{1:L}=(\boldsymbol{x}_1,\boldsymbol{x}_2,...,\boldsymbol{x}_L) \in \mathbb{R}^{L\times C}$ and condition sequence $\boldsymbol{a}^{1:L}=(\boldsymbol{a}_1,\boldsymbol{a}_2,...,\boldsymbol{a}_L) \in \mathbb{R}^{L\times D}$, TiLM stylizes each $\boldsymbol{x}_l$ with the corresponding $\boldsymbol{a}_l$ by regressing the mean and scale factors in FiLM's manner:
\begin{equation}
\label{eq:tilm}
 \boldsymbol{y}_{l}= \boldsymbol{x}_l \odot s(\boldsymbol{a}_l) + m(\boldsymbol{a}_l),
\end{equation}
where ``$\odot$'' denotes the element-wise multiplication and each function of $s$ and $m$ consists of a Mish~\cite{mish} activation and linear layer. 
Compared to Cross-Attention-based conditioning~\cite{facediffuser,codetalker,imitator} with time complexity of $O(L^2)$ for the sequence length $L$, TiLM's time complexity is $O(L)$. 

\subsection{Audio-4D Data Curation with Refinement}
\label{sec:curation}
Because the traditional audio-4D datasets~\cite{voca,multiface} provide limited numbers of videos, 
we extract 3D facial information from 2D videos, similarly to ID-Reveal~\cite{idreveal}.

First, we collect videos from three public datasets, VoxCeleb2~\cite{voxceleb2}, AVSpeech~\cite{avspeech}, and Acappella~\cite{acappella}. 
We filter out inappropriate videos according to the following criteria: 
1) The video duration should be longer than eight seconds, \ie,~sufficient to observe consecutive facial performances for training; 
2) The video should surpass the threshold of $40$ in the image quality assessment score by the HyperIQA~\cite{hyperiqa}.
After the filtering process, we obtain 200,359 videos of eight seconds each and for a total of 445 hours.
Then, we extract 3D information from the collected videos as follows. 
For each set of video frames with a length of $L$ $(=200)$, we initialize the FLAME parameter sequence including shapes $\boldsymbol\alpha^{1:L} \in \mathbb{R}^{L\times|\boldsymbol\alpha|}$,
expressions $\boldsymbol\beta^{1:L} \in \mathbb{R}^{L\times|\boldsymbol\beta|}$, 
and poses $\boldsymbol\gamma^{1:L} \in \mathbb{R}^{L\times|\boldsymbol\gamma|}$ by the pre-trained SPECTRE~\cite{spectre}. 
To disentangle the face shape space from the expression space, we assign the mean shape across frames to the entire video clip. 
We denote it as $\overline{\boldsymbol\alpha}$.
Next, we optimize the parameters 
$\boldsymbol\omega^{1:L}=(\overline{\boldsymbol\alpha}, \boldsymbol\beta^{1:L}, \boldsymbol\gamma^{1:L})$
by the same objectives as SPECTRE but with the single shape:
\begin{equation}
\label{eq:refine}
\hat{\boldsymbol\omega}^{1:L}= \underset{\boldsymbol\omega^{1:L}} {\operatorname{argmin}}~ \mathcal{L}_{em} + \mathcal{L}_{lr} + \mathcal{L}_{c},
\end{equation}
where $\mathcal{L}_{em}$, $\mathcal{L}_{lr}$, and $\mathcal{L}_{c}$ are the emotion loss, lip-reading loss, and geometric constraints, respectively, used in SPECTRE~\cite{spectre}. 
We optimize the parameters for 10 iterations with a learning rate of $5 \times 10^{-5}$ using the Adam~\cite{adam} optimizer. 
More details on our data curation can be found in App.~\ref{sec:curated_dataset_sup}. 

\subsection{Pre-training}
\label{sec:pretraining}
We pre-train our diffusion model on our curated dataset (see Sec.~\ref{sec:curation}). 
In this stage, we train all the parameters but the TCN of Wav2Vec. 
We add noise sequence $\boldsymbol{\epsilon}^{1:L}$ to the clean coefficients $\boldsymbol{z}^{1:L}$ by the diffusion forward process~\cite{ddpm}:
\begin{equation}
\label{eq:diffusion}
\boldsymbol{z}_t^{1:L} = \sqrt{\bar{\alpha}_t} \boldsymbol{z}^{1:L} + \sqrt{1-\bar{\alpha}_t}\boldsymbol{\epsilon}^{1:L},
\end{equation}
where $\bar{\alpha}_t = \prod_{s=1}^{t} 1-\beta_{s}$ and $\beta_{t}$ is a constant noise scale at a diffusion timestep $t$.
Given the noisy coefficients $\boldsymbol{z}_t^{1:L}$, diffusion timestep $t$, and waveform $\boldsymbol{w}^{1:H}$, the model predicts the added noise sequence $\boldsymbol{\epsilon}^{1:L}$. 
We adopt the simple diffusion loss~\cite{ddpm} as follows: 
\begin{equation}
\label{eq:loss_pretraining}
\mathcal{L}_{1}=\mathop{\mathbb{E}}_{\substack{(\boldsymbol{z}^{1:L}, \boldsymbol{w}^{1:H}) \sim \mathcal{D}, \\ t \sim \mathcal{U}[1, T], \\ \boldsymbol{\epsilon}^{1:L}\sim N(\mu_{t},\sigma_{t})}}\left[\lVert \boldsymbol{\epsilon}^{1:L} - \epsilon_{\theta_1}(\boldsymbol{z}_t^{1:L},t,\boldsymbol{w}^{1:H}) \rVert_2^2 \right],
\end{equation}
where $\mathcal{D}$ and $\epsilon_{\theta_1}$ are the pre-training dataset and the denoising network with a weight set $\theta_1$, respectively. 
We denote the optimized $\theta_1$ as $\hat{\theta}_1 = \underset{\theta_1} {\operatorname{argmin}}~\mathcal{L}_{1}$.
It is noteworthy that, differently from the previous reference-assisted approaches~\cite{idreveal,poiforensics,agarwal2020detecting} that perform large-scale pretraining on identity-annotated datasets such as VoxCeleb2~\cite{voxceleb2}, our model is pre-trained only on unlabeled videos, which enables further extension on larger video collections. 

\subsection{Personalization to Specific Subjects}
\label{sec:personalization}
After the pre-training stage, our model works as a strong prior to predict facial expressions from audio.
Here, we leverage it to learn specific talking identities.
Inspired by LLaMA-Adapter~\cite{llamaadapter}, we assign a learnable token sequence $\bm{c}^{1:N} \in \mathbb{R}^{N\times C}$ to each subject.
$\bm{c}^{1:N}$ is projected to be inserted into the self-attention layers by two additional linear layers of $\bm{W}_k \in \mathbb{R}^{C\times C}$ and $\bm{W}_v \in \mathbb{R}^{C\times C}$.
This process is formulated as:
\begin{equation}
\label{eq:gsa}
\begin{aligned}
\bm{h}_i^{1:L}
= \text{Attention}\bigl(\bm{q}_i^{1:L}, [\bm{k}_i^{1:L}, \bm{\tilde{k}}^{1:N}], [\bm{v}_i^{1:L}, \bm{\tilde{v}}^{1:N}]\bigr), \\
\text{where}\quad
\bm{\tilde{k}}^{1:N}
= \bm{W}_k \bm{c}^{1:N}, 
\quad
\bm{\tilde{v}}^{1:N}
= \bm{W}_v \bm{c}^{1:N},
\end{aligned}
\end{equation}
where $\bm{q}_i^{1:L}$, $\bm{k}_i^{1:L}$, and $\bm{v}_i^{1:L}$ are the original query, key, and value sequences of the $i$-th Transformer block, respectively. 
We concatenate $\bm{\tilde{k}}^{1:N}$ and $\bm{\tilde{v}}^{1:N}$ with $\bm{k}_i^{1:L}$ and $\bm{v}_i^{1:L}$, respectively, to insert identity information.
We train only parameters of $\bm{c}^{1:N}$, $\bm{W}_k$, and $\bm{W}_v$ in this stage.
This minimal personalization strategy enables us not only to maintain the original knowledge of the pre-trained model without catastrophic forgetting~\cite{catastrophic}, but also to save spatial memory by not needing to store the base model for different subjects and only storing 528,384 parameters (where $C=512$ and $N=8$) for each subject. 
Considering the face forgery detection task, where the model should be personalized to a lot of subjects, this property is more suitable compared to full fine-tuning.
%
Our objective function is formulated as:
\begin{equation}
\label{eq:loss_personalization}
\mathcal{L}_{2}=\mathop{\mathbb{E}}_{\substack{(\boldsymbol{z}^{1:L}, \boldsymbol{w}^{1:H}) \sim \widetilde{\mathcal{D}}, \\ t \sim \mathcal{U}[1, T], \\ \boldsymbol{\epsilon}^{1:L}\sim N(\mu_{t},\sigma_{t})}}\left[\lVert \boldsymbol{\epsilon}^{1:L} - \epsilon_{\theta}(\boldsymbol{z}_t^{1:L},t,\boldsymbol{w}^{1:H}, \boldsymbol{c}^{1:N}) \rVert_2^2 \right],
\end{equation}
where $\widetilde{\mathcal{D}}$ represents a reference set of a specific subject, and $\theta=\{\hat{\theta}_1,\theta_2\}$ with a weight set $\theta_2=\{\bm{c}^{1:N},\bm{W}_k,\bm{W}_v\}$ for the adapter.
We denote the optimized parameter set as $\hat{\theta}=\{\hat{\theta}_1,\hat{\theta}_2\}$ with the optimized adapter $\hat{\theta}_2 = \underset{\theta_2} {\operatorname{argmin}}~\mathcal{L}_{2}$.

\subsection{Video Authentication with Content-Agnosticity}
\label{sec:authentication}
Once the model is personalized, it can represent subject-specific expression distribution. 
Here, we expose deepfakes with our personalized model by verifying whether the talking identity of an input video is the same as that of the personalized subject.
One of the straightforward approaches to predict the identity distance is to use the reconstruction loss Eq.~\eqref{eq:loss_personalization} as an authentication criterion
as in DiffusionClassifiers~\cite{diffusionclassifier,chen2024robust}, 
where smaller values may be considered closer talking identities to the subject.
However, we found that this approach does not perform well in our case because the values in Eq.~\eqref{eq:loss_personalization} significantly vary depending on their contents, \ie, what and how people talk in videos. 
Therefore, we introduce \textbf{content-agnostic authentication}. 
For each input sequence set of $(\boldsymbol{z}^{1:L}, \boldsymbol{w}^{1:H})$, our authentication score $\mathcal{A}$ is formulated as:

\begin{equation}
\label{eq:authentication}
\mathcal{A}
= \frac{\mathop{\mathbb{E}}_{\substack{t \sim \mathcal{U}[t_{\text{start}}, t_{\text{end}}), \\ \boldsymbol{\epsilon}^{1:L}\sim N(\mu_{t},\sigma_{t})}}
\left\lbrack\lVert \boldsymbol{\epsilon}^{1:L} - \epsilon_{\hat{\theta}}(\boldsymbol{z}_t^{1:L},t,\boldsymbol{w}^{1:H}, \boldsymbol{c}^{1:N}) \rVert_2^2\right\rbrack}{\mathop{\mathbb{E}}_{\substack{t \sim \mathcal{U}[t_{\text{start}}, t_{\text{end}}], \\ \boldsymbol{\epsilon}^{1:L}\sim N(\mu_{t},\sigma_{t})}}
\left\lbrack\lVert \boldsymbol{\epsilon}^{1:L} - \epsilon_{\hat{\theta}_1}(\boldsymbol{z}_t^{1:L},t,\boldsymbol{w}^{1:H}) \rVert_2^2\right\rbrack},
\end{equation} 
where the denominator works as an adaptive scaling value to cancel the variances of the video contents, which enables evaluating the purer identity distance.
Intuitively, when the identity of an input video is the same as the personalized subject, the numerator is smaller than the denominator. 
A notable point in Eq.~\eqref{eq:authentication} is which diffusion timesteps $t$ should be sampled for authentication. 
We observe that removing timesteps at both sides, \eg, from 1 to 200 and from 801 to 1000, improves the discriminability of our model. 
This is because 1) at small timesteps, the data is minimally perturbed by noise, rendering the noise prediction task trivially easy as the original data structure remains largely intact~\cite{ddpm}, and 2) at large timesteps, the data is almost entirely dominated by noise, making the prediction excessively challenging~\cite{ddim}. 
These extreme situations blur the difference between the numerator and denominator in Eq.~\eqref{eq:authentication}, \ie, prediction errors with and without identity conditioning. 
Therefore, we empirically set $t_{\text{start}}$ and $t_{\text{end}}$ in Eq.~\eqref{eq:authentication} to 201 and 800, respectively, which is explored in Table~\ref{tb:ablation_sampling} in the supplementary material. 
We also discuss the effect of different numbers of sampled noise sequences $\bm{\epsilon}^{1:L}$ in Fig.~\ref{fig:comprehensive_analysis}(b). 
For thresholding to make real/fake decision on our unbounded authentication score, please refer to App.~\ref{sec:additional_results}.

\subsection{Implementation Details}
We implement our model in PyTorch~\cite{paszke2019pytorch}. 
All the experiments are conducted with a single NVIDIA A100 GPU.
For pre-training, we optimize our model for $100$ epochs.
We use the ADAN optimizer~\cite{adan}, and the batch size and learning rate are set to 256 and $10^{-4}$, respectively. 
Pre-training finishes within a day.
For personalization, we extract expression parameters from reference sets in a similar manner to Eq.~\eqref{eq:refine} but use single shape parameters shared across videos belonging to the same subjects.
We insert eight adapter tokens for each subject.
Because the reference sets are relatively small, we over-sample the videos by the batch size (=256), which enables us to personalize our model on a large batch size for stabilized diffusion training. 
As a result, the number of iterations for each subject is (\#videos)$\times$(\#epochs), where we set the number of epochs to 100.
Personalization finishes within 15 minutes for each subject with eight videos. 
During authentication, the diffusion timesteps $t$ are sampled at 60 equally spaced points in the range $[201,800]$, and on each $t$, we sample 64 different noise sequences $\boldsymbol{\epsilon}^{1:L}$.
Inference, excluding feature extraction of $\boldsymbol{z}^{1:L}$ and $\boldsymbol{w}^{1:H}$, for a single video with eight seconds takes 25 seconds. 
More details can be found in App.~\ref{sec:imple_ours_sup}

\begin{table*}[t]
    \small
    \centering
    \begin{adjustbox}{width=0.95\textwidth}
        \begin{tabular}{ccccccccc|c} \toprule
        \multirow{2}{*}{Learning Type}&\multirow{2}{*}{Method} & \multirow{2}{*}{Venue}& \multirow{2}{*}{Modality}&\multirow{2}{*}{Reference}& \multicolumn{5}{c}{Test Set AUC (\%)}\\ 
        \cmidrule(lr){6-10}
        &&& &  & DF-TIMIT & DFDCP & KoDF & IDForge  & \textbf{Avg}\\
        \midrule
        \multirow{11}{*}{Supervised} & EfficientNet-b4~\cite{efficientnet}&ICML'19     & Frame          &  & 94.67 & 60.15 & 70.21 & 75.88 & 75.23\\
        &LipForensics~\cite{lipforensics} &CVPR'21    & Video          &  & 96.74 & 69.89 & 95.98 & 93.05 & 88.92 \\
        &FTCN~\cite{ftcn}   &ICCV'21        & Video          &   & \textbf{99.91} & 61.19 & 86.06 & 93.52 & 85.17\\
        &RECCE~\cite{recce}&CVPR'22 & Frame && 68.42 & 49.76 & 53.13 & 56.19 & 56.88\\
        &RealForensics~\cite{realforensics}&CVPR'22     & Video          &  & 97.74 & 72.23 & 72.56 & 78.19 & 80.18 \\
        &AltFreezing~\cite{altfreezing} &CVPR'23 & Video          &  & 99.82 & 69.33 & 96.41 & \textbf{95.79} & 90.34\\
        &UCF~\cite{ucf}  &ICCV'23    & Frame          &   & 89.19 & 77.10 & 62.71 & 81.46 & 77.12 \\
        &LipFD~\cite{liu2024lips}&NeurIPS'24  & Audio \& Video & & 54.78 & 56.13 & 50.76 & 64.33 & 56.50 \\
        &FSFM~\cite{fsfm} &CVPR'25             & Frame & & 90.56 &86.94&74.67&87.59 & 84.94\\
        & DFD-FCG~\cite{dfdfcg} & CVPR'25 & Video & & 99.02 & 90.04 & \textbf{97.40} & 77.56 & \textbf{91.00} \\
        & EFFORT~\cite{effort} & ICML'25 &Frame & & 94.96& \textbf{92.89}&85.22 &85.79& 89.72\\
        \midrule
        \multirow{5}{*}{Pseudo-Supervised} & Face X-ray~\cite{facexray} &CVPR'20     & Frame          &   & 76.01 & 67.77 & 48.84 & 54.76 & 61.35 \\
        &SBI~\cite{sbi} &CVPR'22           & Frame          &   & 84.71 & 88.51 & \textbf{87.84} & 80.94 & 85.50\\
        &ICT~\cite{ict}   &CVPR'22              & Frame          &            & 77.35 & 74.59 & 50.53 & 59.46 & 65.48 \\
        &LAA-Net w/ SBI~\cite{laanet}&CVPR'24               & Frame          &  & 78.51 & 88.96 & 82.25 & 79.90 & 82.41 \\
        &ForensicsAdapter~\cite{forensicsadapter} &CVPR'25 & Frame & & \textbf{97.32} & \textbf{93.98} & 86.10 & \textbf{83.92} & \textbf{90.33}\\
        \midrule
        \multirow{6}{*}{Self-Supervised} & ID-Reveal~\cite{idreveal} &ICCV'21      & Video          & \checkmark  & 60.41 & 83.84 & 61.64 &74.77 &70.17\\
        &AVAD~\cite{avad}  &CVPR'23              & Audio \& Video &   & 77.39 & 45.58 & 61.41 & 55.78 & 60.04 \\
        &POI-Forensics~\cite{poiforensics}&CVPRW'23       & Audio \& Video & \checkmark & 85.65 &85.34 & 64.91& 72.60 & 77.13 \\
        &SpeechForensics~\cite{speechforensics}& NeurIPS'24  & Audio \& Video & &71.35 & 63.61 & 82.98 & \textbf{93.66} & 77.90\\
        &\cellcolor[gray]{0.9}\textbf{Ours} w/ VoxCeleb2&\cellcolor[gray]{0.9}-          & \cellcolor[gray]{0.9}Audio \& Video & \cellcolor[gray]{0.9}\checkmark  &\cellcolor[gray]{0.9}99.49 & \cellcolor[gray]{0.9}93.06 &\cellcolor[gray]{0.9} 87.73 &\cellcolor[gray]{0.9} 93.25 &\cellcolor[gray]{0.9} 93.38 \\
        &\cellcolor[gray]{0.9}\textbf{Ours} &\cellcolor[gray]{0.9}-          & \cellcolor[gray]{0.9}Audio \& Video & \cellcolor[gray]{0.9}\checkmark  &\cellcolor[gray]{0.9}\textbf{99.72} &\cellcolor[gray]{0.9} \textbf{93.45} &\cellcolor[gray]{0.9} \textbf{95.31} &\cellcolor[gray]{0.9} 92.40 &\cellcolor[gray]{0.9} \textbf{95.22} \\
        \bottomrule
        \end{tabular}
    \end{adjustbox}
  \caption{\textbf{Generalization ability to unseen forgeries on DF-TIMIT, DFDCP, KoDF, and IDForge datasets.} The best value for each learning type on each test set is highlighted in \textbf{bold}. 
  Our method achieves 95.22\% in terms of the average AUC, resulting in ${\approx}5\%$ increase over the previous best methods DFD-FCG and ForensicsAdapter (CVPR'25).
  } 
  \label{tb:cross_dataset}
\end{table*}

\section{Experiments}

\subsection{Problem Setting}
\label{sec:problem} 
The objective of face forgery detection is to determine whether a given video contains manipulated facial content. 
We focus on the \textit{person-of-interest} scenario introduced by~Agarwal~\etal~\cite{agarwal2019protecting},
where the detector can access a set of pristine reference videos per subject. 
This setting is practical in real-world applications, as reference videos can often be
collected online. 
%
Thanks to the proliferation of social media platforms, obtaining such reference videos has become increasingly feasible, also for non-experts. 
%

\subsection{Setup}
\label{sec:setup}
\noindent\textbf{Baselines.}
We refer to 20 state-of-the-art methods that can be categorized into three groups based on the learning types:
1) \textit{Supervised methods}~\cite{efficientnet,lipforensics,ftcn,recce,realforensics,altfreezing,ucf,liu2024lips,fsfm,effort,dfdfcg} are trained on actual fake samples from FaceForensics++ (FF++)~\cite{ffpp} except one~\cite{liu2024lips} trained on Wav2Lip-modified LRS3~\cite{liu2024lips}.
2) \textit{Pseudo-supervised methods}~\cite{facexray,ict,sbi,laanet,forensicsadapter}\footnote{We do not adopt ICT-Ref~\cite{ict} because it assumes dataset-level reference sets, which is totally different from our problem setting.} are trained on synthetic fake samples generated to cover a variety of artifacts seen in deepfakes.
3) \textit{Self-supervised methods}~\cite{avad,idreveal,poiforensics,speechforensics} are trained without any fake samples, aiming at independence from overfitting to specific forgery patterns.
All the models are reproduced using official or third-party implementations except EfficientNet-b4 and Face X-ray, which we re-implement. 
We also adopt our model pre-trained only on VoxCeleb2~\cite{voxceleb2} to compare our method with the previous methods~\cite{idreveal,poiforensics} on the same training set.
More details can be found in App.~\ref{sec:imple_previous_sup}. 

\noindent\textbf{Evaluation Datasets.}
We refer to four audio-visual deepfake detection datasets for evaluation, including
Deepfake-TIMIT (DF-TIMIT)~\cite{dftimit}, Deepfake Detection Challenge Preview (DFDCP)~\cite{dfdcp}, Korean Deepfake Detection (KoDF)~\cite{kodf}, and Identity-Driven Multimedia Forgery Detection (IDForge)~\cite{idforge} datasets, where the numbers of subjects are 32, 39, 67, and 53, respectively.
Note that we do not adopt some conventional datasets due to the absence of audio channels~\cite{ffpp,cdf,cdfpp,dfd,ffiw,wilddf} and identity annotations~\cite{dfdc}.
Also, we do not adopt the datasets~\cite{favc,lavdf,avdf1m} built on VoxCeleb2~\cite{voxceleb2} because of the overlap with the training data of our method, ID-Reveal, and POI-Forensics.
%
See App.~\ref{sec:evaluation_dataset_construction} for further details on the evaluation datasets. 
%


\noindent\textbf{Sora2 Cameo Forensics Preview (S2CFP) Dataset.}
To evaluate the model generality not only on traditional deepfake datasets but also on the most recent video generative model Sora2~\cite{sora} with  Cameo (\ie, personalized video generation) feature, we introduce a new dataset, called Sora2 Cameo Forensics Preview (S2CFP) dataset which serves as a starting point for face forensics on Sora2. 
The dataset provides, for each subject, a reference set that consists of 40 real videos and a test set that consists of 12 real videos and 12 fake ones. 
For more details, please refer to App.~\ref{sec:evaluation_dataset_construction}.

\noindent\textbf{Evaluation Metric.}
 We adopt the video-level area under the ROC curve (AUC).
%
For the frame-level methods, we follow the official or third-party inference strategies. 
If there is no instruction, we simply average the predictions over frames. 
We also report the average AUC denoted as ``Avg'' to evaluate the generalization ability across the datasets. 
%
%

\subsection{Comparison with Previous Methods}
Table~\ref{tb:cross_dataset} shows the generalization ability to unseen forgeries. 
We also report for each method the learning type, input modality, and usage of reference. 
Overall, our method achieves an average AUC of 95.22\% on DF-TIMIT, DFDCP, KoDF, and IDForge, demonstrating the highest generalization ability over datasets. 
The supervised methods often suffer from the domain gap between training and test sets (\eg, 69.33 by AltFreezing on DFDCP and 77.56 by DFD-FCG on IDForge). 
The pseudo-supervised methods, such as SBI~\cite{sbi} and ForensicsAdapter~\cite{forensicsadapter} stabilize detection performance by pseudo-fake augmentation; however, they fail to generalize on KoDF and IDForge datasets.
The previous self-supervised approaches~\cite{avad,idreveal,poiforensics} struggle to learn effective features only from self-supervision, resulting in poor performance, \eg, the average AUC of 77.13 by POI-Forensics. 
The result indicates that our method suggests a promising research direction for addressing the generalization problem only with self-supervision.

Moreover, our model trained only on VoxCeleb2~\cite{voxceleb2} still outperforms the previous state-of-the-art methods. 
We can see that this variant drops performance on KoDF from 95.31 to 87.73 because KoDF focuses on the Korean language, which hardly appears in VoxCeleb2; our additional curated training data, especially AVSpeech~\cite{avspeech} that includes a variety of languages, improves AUC on non-English videos.

\begin{table}[t]
    \centering
    \begin{adjustbox}{width=1.0\linewidth}
    \begin{tabular}{ccccc|c} \toprule
    \multirow{2}{*}{Detector Type}&\multirow{2}{*}{Method}& \multicolumn{4}{c}{Test Set AUC (\%) on Each Subject}\\ 
        \cmidrule(lr){3-6}
      & &@ijustine &@mcuban & @sama & \textbf{Avg}\\
      \midrule
        \multirow{8}{*}{\shortstack{Face Forgery \\Detector}}&LipForensics & 48.61 & 56.94 & 33.33 & 46.29\\
         &AltFreezing & 27.78 & 38.19 & 15.97 & 27.31\\
        &DFD-FCG & 36.11 & 53.47 & 56.25 & 48.61\\
        &EFFORT & 85.42 & 56.94 & 61.03 & 67.80\\
        &ForensicsAdapter & 38.89 & 82.64 & 62.50 & 61.34\\
        &POI-Forensics & 41.67 & 43.75 & 72.22 & 52.55\\
        &SpeechForensics & 54.86 & 64.58 & 63.89 & 61.11\\
        &\cellcolor[gray]{0.9}\textbf{Ours} & \cellcolor[gray]{0.9}\textbf{98.61} & \cellcolor[gray]{0.9}\textbf{84.72} & \cellcolor[gray]{0.9}\textbf{100.00} & \cellcolor[gray]{0.9}\textbf{94.44}\\
        \midrule
        \multirow{3}{*}{\shortstack{Diffusion-Generated \\Image Detector}} & DIRE & 11.11 & 56.25 & 43.75 & 37.04\\
         & AEROBLADE &43.75 &52.08 &91.67 &62.50 \\
         & B-Free & 65.97 & 71.53 & 76.39 & 71.30 \\
      \bottomrule
    \end{tabular}
    \end{adjustbox}
  \caption{\textbf{Detection capability on Sora2-generated videos.}}
  \label{tb:sora}
\end{table}

\subsection{Comparison on Sora2-Generated Videos}
Here, we evaluate the detectors on our S2CFP dataset described in Sec.~\ref{sec:setup}.
We give the result on three subjects, \ie, @ijustine, @mcuban, and @sama in Table~\ref{tb:sora} with stronger or more important baselines from Table~\ref{tb:cross_dataset}.
In addition, we refer to the state-of-the-art diffusion-generated image detectors~\cite{dire,aeroblade,bfree} for further comparison.
We observe that the previous methods of both detector types fail to generalize to Sora2, while our method achieves the average AUC of 94.44\%, demonstrating the generalization ability of our model.
The comprehensive result is found in App.~\ref{sec:additional_results}.

\subsection{Robustness to Perturbations}
\begin{figure}[t]
  \centering
  \vspace{-3.0mm}
  \begin{adjustbox}{width=1.0\linewidth}
  \includegraphics{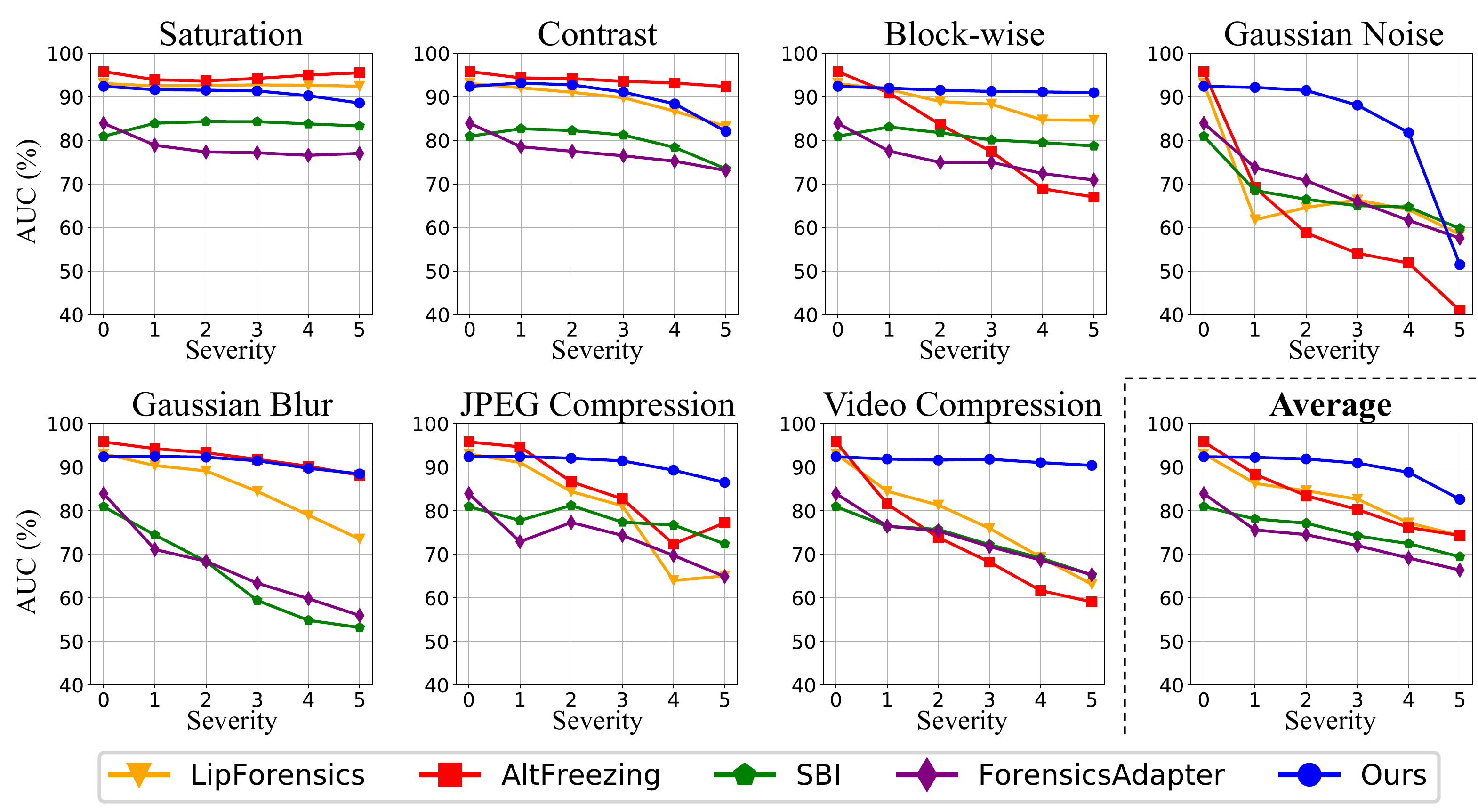}
  \end{adjustbox}
  \caption{\textbf{Robustness to common corruptions on IDForge.} Severity levels are defined in DeeperForensics~\cite{deeperforensics}. Our method is highly consistent on the perturbations especially compression that detectors encounter frequently in real-world scenarios.} 
  \label{fig:robustness}
\end{figure}

Next, we evaluate models on corrupted videos, which detectors sometimes encounter in real-world scenarios.
Following DeeperForensics~\cite{deeperforensics}, we adopt seven perturbations, including color saturation, color contrast, block-wise, Gaussian noise, Gaussian blur, JPEG compression, and video compression with five different severity levels \{1,2,3,4,5\}. 
In Fig.~\ref{fig:robustness} where ``severity $=0$'' means no corruption, we give the result of our method in comparison with LipForensics, AltFreezing, SBI, and ForensicsAdapter that achieve better generalization than the other previous approaches as shown in Table~\ref{tb:cross_dataset}, except DFD-FCG and EFFORT, which we do not adopt because of the insufficient AUC compared to the other supervised methods on IDForge.
Our method is highly robust to the corruptions, especially block-wise, Gaussian blur and compression.
We observe in the ``Average'' figure that although LipForensics and AltFreezing slightly perform better than ours when no corruption is added, they immediately drop performance when severity $>0$. 
In contrast, our method maintains the consistent AUC, demonstrating the applicability in real-world scenarios. 

\begin{figure}[t]
  \centering
  \begin{adjustbox}{width=1.0\linewidth}
  \begin{minipage}{.33\linewidth}
    \centering
    \includegraphics[height=2.15cm]{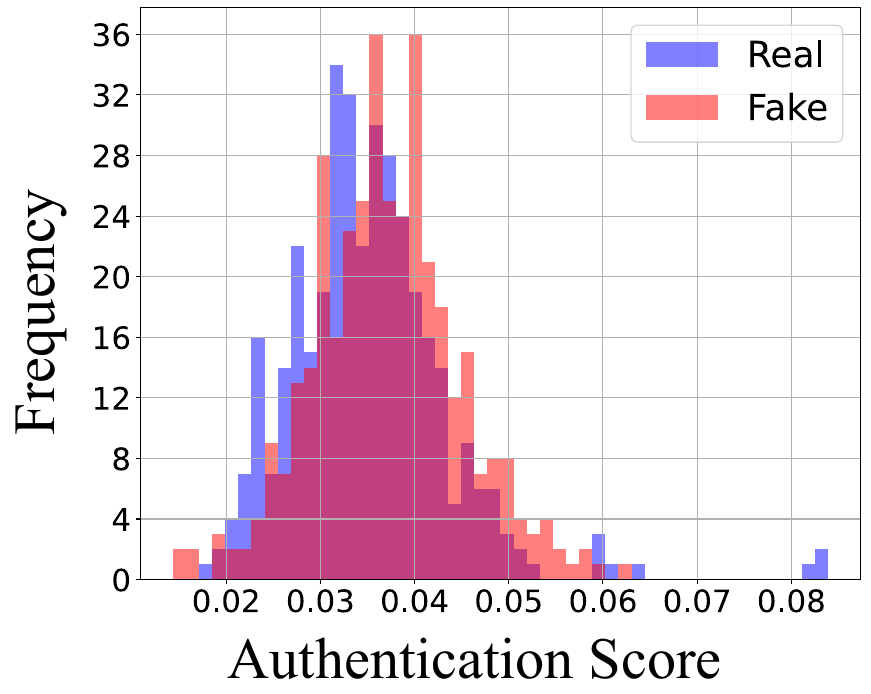} 
    \subcaption{$d_{1}$ (AUC = 58.88)}
    \label{fig:baseline_dt}
  \end{minipage}
  \begin{minipage}{.33\linewidth}
    \centering
    \includegraphics[height=2.15cm]{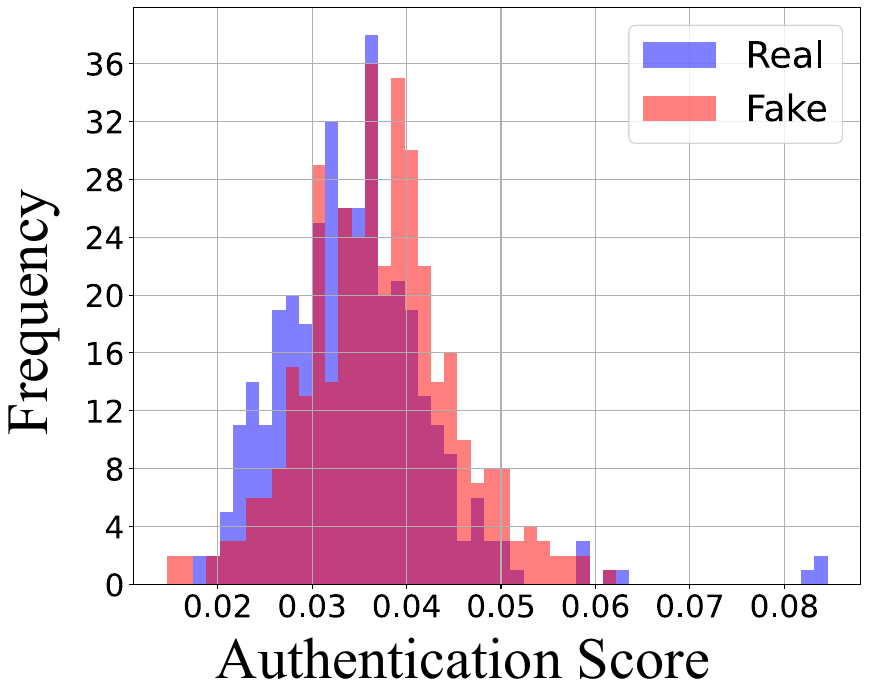}
    \subcaption{$d_{2}$ (AUC = 61.34)}
    \label{fig:baseline_dp}
  \end{minipage}
  \begin{minipage}{.33\linewidth}
    \centering
    \includegraphics[height=2.15cm]{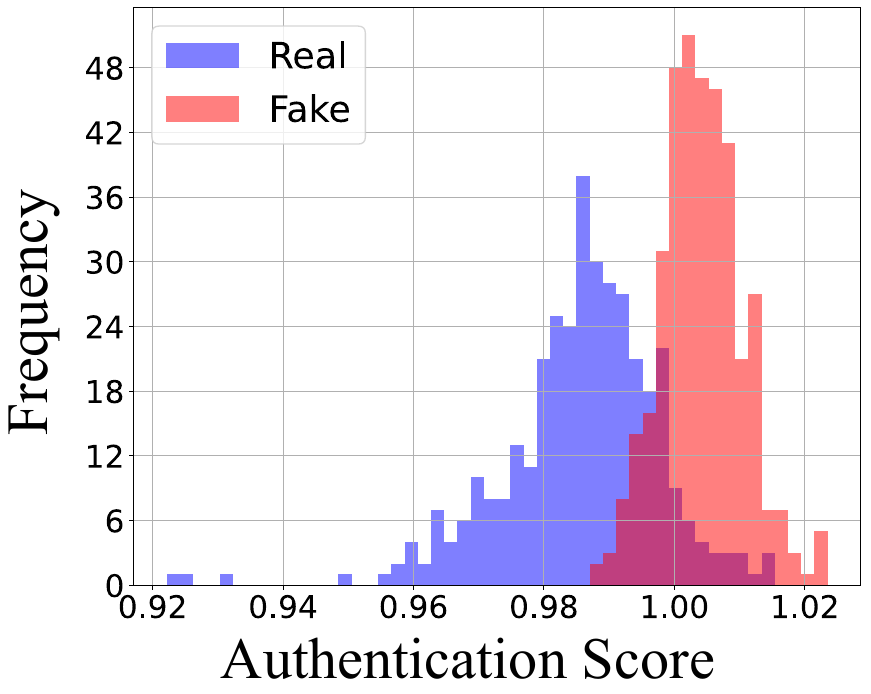}
    \subcaption{Ours (AUC = 93.45)}
    \label{fig:our_criterion}
  \end{minipage}
  \end{adjustbox}
  \caption{\textbf{Effect of the content-agnostic authentication.} The direct use of objectives for (a) pre-training and (b) personalization does not work for authentication, while (c) their quotient significantly distinguishes fake samples from real ones.}
  \label{fig:ablation_criterion}
\end{figure}

\subsection{Framework Analysis} 
\noindent\textbf{3DMM Extraction Strategy.}
We investigate the effect of our 3DMM extraction strategy, including feed-forward initialization by SPECTRE~\cite{spectre} and refinement processes described in Sec.~\ref{sec:model}.
We compare our model with a variance trained on expression sequences directly extracted by SPECTRE without our refinement process.
We observe that the model without refinement results in 46.82\% in AUC on the DFDCP dataset.
This is because SPECTRE predicts different face shape parameters for different frames; there is a crucial entanglement between face shape and expression. 
In contrast, our refinement process, which singularizes the face shape and iteratively optimizes the error in Eq.~\eqref{eq:refine}, disentangles expression parameters from the face shape parameter and enables our model to distinguish real expressions from fake ones, achieving 93.45\% on DFDCP. 

\noindent\textbf{Authentication Strategy.}
We compare our content-agnostic authentication  Eq.~\eqref{eq:authentication} with two other possibilities in Fig.~\ref{fig:ablation_criterion}: 
(a) the distance used in pre-training, \ie, $d_{1} = \lVert \boldsymbol{\epsilon}^{1:L} - \epsilon_{\hat{\theta}_1}(\boldsymbol{z}_t^{1:L},t,\boldsymbol{w}^{1:H}) \rVert_2^2$ and
(b) one in personalization, \ie, $d_{2} = \lVert \boldsymbol{\epsilon}^{1:L} - \epsilon_{\hat{\theta}}(\boldsymbol{z}_t^{1:L},t,\boldsymbol{w}^{1:H}, \boldsymbol{c}^{1:N}) \rVert_2^2$. 
As shown in the figure, the direct use of the objectives (Figs.~\ref{fig:ablation_criterion}(a) and \ref{fig:ablation_criterion}(b)) fails to distinguish fake samples from real ones, while our content-agnostic authentication works as a strong indicator for face forgery detection (Fig.~\ref{fig:ablation_criterion}(c)). 

\begin{figure}[t]
  \centering
  \begin{adjustbox}{width=1.0\linewidth}
  \includegraphics{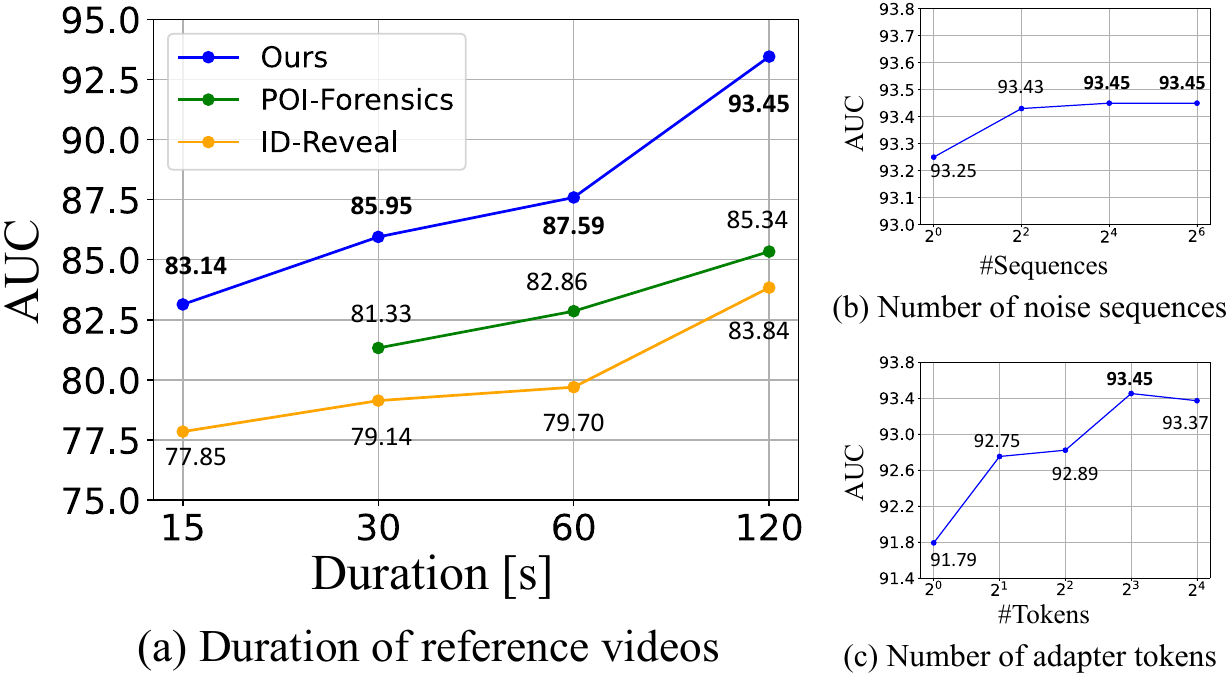}
  \end{adjustbox}
  \caption{\textbf{Analyses on \methodname.} (a) Our model achieves a higher AUC on different durations of reference videos than the previous methods. (b) Detection accuracy converges well with 64 noise sequences. (c) The adapter with eight tokens per subject brings the most accurate results.} 
  \label{fig:comprehensive_analysis}
\end{figure}

\noindent\textbf{Duration of Reference Videos.} 
It is important to examine the effect of the size of the reference set for personalization.
Here, we personalize our model on different reference video durations, $\{15,30,60,120\}$; see  Fig.~\ref{fig:comprehensive_analysis}(a). 
We also evaluate the previous reference-assisted approaches, ID-Reveal~\cite{idreveal} and POI-Forensics~\cite{poiforensics}. 
Note that we could not obtain the result of POI-Forensics on a duration of 15 seconds corresponding to a single reference video because the method requires at least two videos to normalize predicted values.
We see that our method outperforms the previous methods in all cases with consistent margins.
Moreover, given longer durations, the result of our model greatly improves, while those of the previous approaches show marginal improvements. 
The result also implies that our method could be further improved with additional reference videos.

\noindent\textbf{Number of Sampled Noise Sequences.}
Our model can better estimate the authentication score by sampling more noise sequences, as can be seen from Eq.~\eqref{eq:authentication}. 
Hence, we investigate the effect of the number of sampled sequences on the model's performance. 
We plot the results with different numbers of noise sequences, \ie, $\{1,4,16,64\}$, in Fig.~\ref{fig:comprehensive_analysis}(b). 
Though more noise slightly improves the result, our model maintains the performance even with a single noise sequence compared to 64 sequences (93.25\% vs 93.45\%), which shows the robustness of our method to the randomness of added noise sequences.
We observe that the authentication values converge well with 64 sequences. 

\begin{figure}[t]
  \centering
  \begin{adjustbox}{width=1.0\linewidth}
  \includegraphics{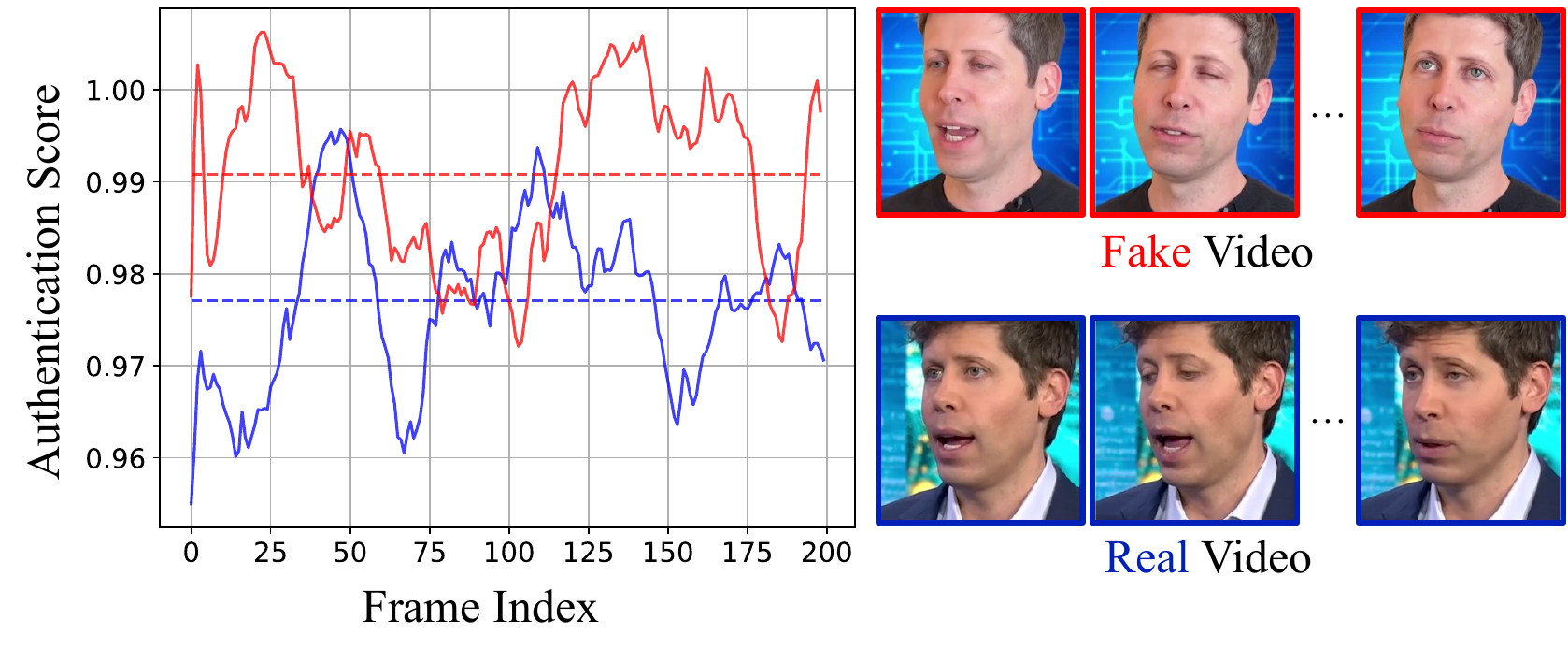}
  \end{adjustbox}
  \vspace{-5mm}
  \caption{\textbf{Temporal visualization of the authentication score.} The solid blue and red lines represent the authentication scores over the frame index of a real video and a Sora2-generated video mimicking the subject, respectively. The dotted lines are the averaged values, which are statistically lower for real videos than for the corresponding fake videos.} 
  \label{fig:visualization}
\end{figure}

\noindent\textbf{Number of Adapter Tokens.}
We also explore the optimal number of adapter tokens inserted for personalization.
We show the results with different numbers of tokens, \ie, $\{1,2,4,8, 16\}$, in Fig.~\ref{fig:comprehensive_analysis}(c). 
We can see that adding some tokens improves the results (from 91.79\% with a single token to 93.45\% with eight tokens); however, excessive tokens slightly worsen them due to over-fitting to reference data (from 93.45\% with eight tokens to 93.37\% with 16 tokens). 
We recommend using eight tokens by default.
\noindent\textbf{Temporal Visualization.} 
In Fig.~\ref{fig:visualization}, we visualize our authentication score on a real video and the corresponding Sora2-generated video of SCFP by unfolding Eq.~\ref{eq:authentication} in the temporal dimension. 
We average each point with a window size of 15 for better viewability and present their average values as dotted lines. 
We can see that the scores for the real video are lower than those for the fake one, indicating that our model is capable of distinguishing real and fake videos.
More examples are included in App.~\ref{sec:additional_results}. 

\section{Limitations} 
We observe some limitations:
1) Our model depends on the capabilities of the off-the-shelf models, \ie, FLAME~\cite{flame} for face representation, SPECTRE~\cite{spectre} for feed-forward initialization, and Wav2Vec 2.0~\cite{wav2vec} for audio encoding. 
2) Our model has computational overhead during inference mainly due to the iterative optimization for 3DMM extraction and diffusion reconstruction over multiple timesteps and noise sequences, which we discuss further in App.~\ref{sec:additional_results}.

\section{Conclusion} 
We present \methodname, a self-supervised face forgery detection framework based on an audio-to-expression diffusion model. 
Our most important observation is that personalizing audio-to-expression diffusion models to specific subjects enables person-of-interest face forgery detection without any prior knowledge of face forgeries. 
Extensive experiments demonstrate that our approach significantly outperforms the previous state-of-the-art methods in generalization to unseen manipulations, including not only traditional deepfakes but also Sora2 and in robustness to common corruptions, suggesting a promising research direction for real-world face forgery detection.
\appendix 

\clearpage
\maketitlesupplementary
In this document, we provide information about (A) the evaluation data setup, (B) the curated dataset for pre-training, (C) the implementation details of our model and baselines, and (D) additional experimental results.

\section{Evaluation Dataset}
\label{sec:evaluation_dataset_construction}
Most existing deepfake datasets do not provide reference sets explicitly. Therefore, we carefully construct evaluation sets for the experiments.
To compare our model with previous methods, datasets should satisfy the following requirements: datasets should include (1) an audio channel, (2) identity labels, and (3) multiple clips for each subject. 
(4) We also exclude datasets~\cite{favc,lavdf,avdf1m} built on VoxCeleb2~\cite{voxceleb2} because some methods~\cite{idreveal,poiforensics} and ours are trained on the dataset.
We report the attributes of existing deepfake video datasets in Table~\ref{tb:deepfake_datasets}.
Through the evaluation datasets, we exclude videos that are not well face-tracked from our experiments, following the convention (\eg, seen in the LipForensics paper~\cite{lipforensics}).
We describe the construction of our evaluation datasets below:

\noindent\textbf{DF-TIMIT~\cite{dftimit}.} 
This dataset is built on the VidTIMIT~\cite{Sanderson2002TheVD} dataset. 
It includes short-term videos of 32 subjects. 
Each subject has 10 videos with different sentences.
They are split into three sessions according to when they were recorded. 
We use six videos of Session 1 for reference and four remaining videos of Sessions 2 and 3 for testing.
Also, though this dataset provides two different qualities of deepfake videos, we adopt the higher (\ie, more challenging) one as the lower one is easy to detect for current state-of-the-art methods.

\noindent\textbf{DFDCP~\cite{dfdcp}.} 
This dataset was released as a preview for a Kaggle competition on deepfake detection\footnote{\url{https://www.kaggle.com/competitions/deepfake-detection-challenge}}. 
It contains two types of face-swapped deepfake videos.
We uniformly sample eight videos for reference and use the remaining videos for evaluation, ensuring that the same scene indexes do not overlap between the reference and test sets.
The evaluation set contains 39 subjects, and its test set consists of 128 real and fake videos.
We uniformly sample real and fake videos for the test set, keeping the ratio of real/fake as close to 1 as possible.

\begin{table}[t]
    \centering
    \begin{adjustbox}{width=1.0\linewidth}
    \begin{tabular}{l c c c c}
        \toprule
        Dataset & \makecell{Audio\\channel} & \makecell{Identity\\label} & \makecell{Multiple clips\\for each subject}  & \makecell{Independent from\\VoxCeleb2} \\
        \midrule
        \rowcolor[gray]{0.9} DF-TIMIT~\cite{dftimit} & \checkmark & \checkmark & \checkmark &\checkmark \\
        UADFV~\cite{uadfv}&  &  & &\checkmark  \\
        FF++~\cite{ffpp} &  &  & &\checkmark  \\
        DFD~\cite{dfd} &  & \checkmark & \checkmark & \checkmark  \\
        Celeb-DFv2~\cite{cdf}  &  & \checkmark & \checkmark &\checkmark\\
        \rowcolor[gray]{0.9} DFDCP~\cite{dfdcp} & \checkmark & \checkmark & \checkmark &  \checkmark\\
        DFDC~\cite{dfdc} &  \checkmark &  &  & \checkmark \\
        DeeperForensics~\cite{deeperforensics}&   &  &  & \checkmark \\
        FFIW~\cite{ffiw} &  &  &&\checkmark  \\
        \rowcolor[gray]{0.9} KoDF~\cite{kodf} & \checkmark & \checkmark & \checkmark & \checkmark\\
        FakeAVCeleb~\cite{favc} & \checkmark & \checkmark & \checkmark &   \\
        LAV-DF~\cite{lavdf} & \checkmark & \checkmark & \checkmark & \\
        AV-Deepfake1M~\cite{avdf1m} & \checkmark & \checkmark  & \checkmark \\
        \rowcolor[gray]{0.9} IDForge~\cite{idforge} &  \checkmark & \checkmark & \checkmark & \checkmark \\
        Celeb-DF++~\cite{cdfpp}  &  & \checkmark & \checkmark &\checkmark\\
        \bottomrule
    \end{tabular}
    \end{adjustbox}
    \caption{\textbf{Attributes of deepfake video datasets.} In the main paper, we adopt DF-TIMIT, DFDCP, KoDF, and IDForge datasets that satisfy the requirements for fair evaluation.}
    \label{tb:deepfake_datasets}
\end{table}

\begin{table}[t]
    \centering
    \begin{adjustbox}{width=1.0\linewidth}
    \begin{tabular}{lcccccc}
        \toprule
        \multirow{2}{*}{Dataset} & \multirow{2}{*}{\#ID}&\multicolumn{2}{c}{Reference Set} & \multicolumn{3}{c}{Test Set}\\ 
        \cmidrule(lr){3-4}\cmidrule(lr){5-7}
        & & \#Real & Duration[s] & \#Real & \#Fake & Duration[s] \\
        \midrule
        DF-TIMIT    &32     &192    & 4.42  & 128 & 128 & 4.00\\
        DFDCP       &39     &312    & 15.08 & 366 & 378 & 8.00\\
        KoDF        &67     &4280   & 8.00  & 268 & 268 & 8.00\\
        IDForge     &53     &2880   & 6.89  & 187 & 184 & 7.08\\
        S2CFP        &3      &120     & 8.00  & 36  & 36  & 8.00\\
        \bottomrule
    \end{tabular}
    \end{adjustbox}
    \caption{\textbf{Statistics of our evaluation datasets.} \#ID, \#Real, \#Fake, and Duration represent the number of identities, the number of real and fake videos, and the mean duration (in seconds) per video, respectively.}
    \label{tb:evaluation_datasets}
\end{table}

\noindent\textbf{KoDF~\cite{kodf}.}
This dataset focuses on the Korean language.
We exclude face-swapped videos (\ie, DeepFake-FaceSwap~\cite{deepfake-faceswap}, DeepFaceLab~\cite{dfl}, and FSGAN~\cite{fsgan}) and only adopt face-reenacted videos (\ie, Audio-driven methods~\cite{wav2lip,audiodriven} and First Order Motion Model~\cite{fomm}) for evaluation.
This is because this dataset includes only a single scene (\eg, background and cloth) for each subject; reference-assisted methods can easily spot face-swapped videos by just comparing such irrelevant attributes of input videos with those of reference ones, not based on the talking identities.
The evaluation set includes 67 subjects.
We use 64 videos at most for each reference set with a duration of eight seconds.
We uniformly sample at most four respective real and fake videos for each subject; we exclude videos whose faces are not tracked well, and exclude subjects who do not have one or more well-tracked videos for real and fake classes.

\begin{figure*}[t]
  \centering
  \includegraphics[width=1.0\linewidth]{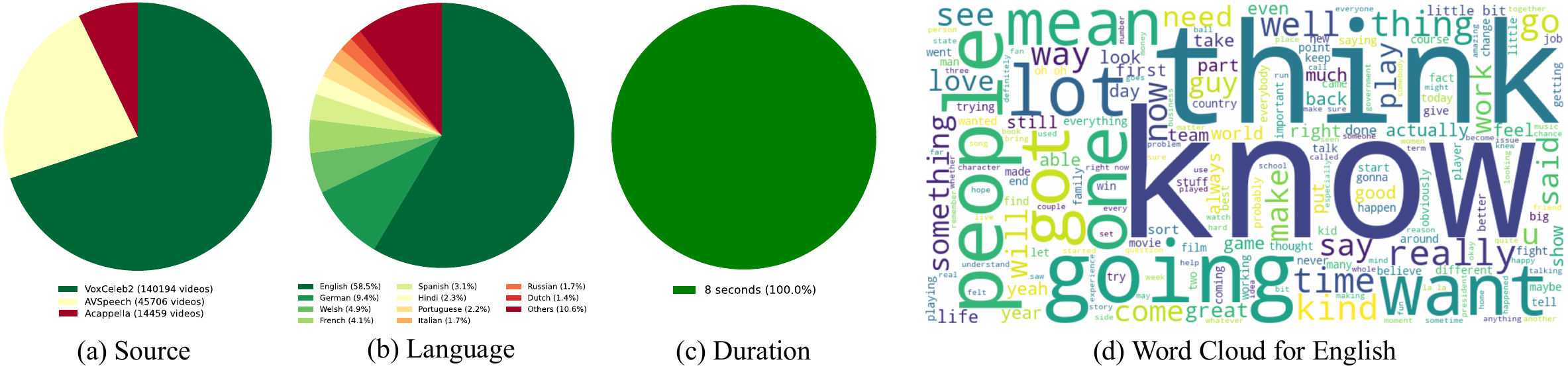}
  \caption{
  \textbf{Pre-training dataset statistics.} 
   (a) We collect monocular videos from three sources: VoxCeleb2, AVSpeech, and Acappella datasets.
   (b) We visualize the ratio of languages spoken in our dataset. 
   (c) All the sequences of our dataset are clipped into eight seconds. 
   (d) We also visualize the Word Cloud for English.
  }
  \label{fig:pretrain_statistics}
\end{figure*}

\noindent\textbf{IDForge~\cite{idforge}.} 
This dataset provides a large-scale video collection for reference-assisted face forgery detection. 
It pre-defines the reference and test sets for each subject; therefore, we follow the official split.
The evaluation set includes 67 subjects.
We use 64 videos at most for each reference set, with a duration of eight seconds at most. 
We uniformly sample at most four respective real and fake videos for each subject; we exclude videos whose faces are not tracked well, and exclude subjects who do not have one or more well-tracked videos for real and fake classes.

\noindent\textbf{S2CFP.}
We create a new dataset to evaluate detection models on the most recent video generation model Sora2~\cite{sora}.
It provides the Cameo feature that enables us to generate specific subjects, although Sora2 does not allow us to generate specific subjects directly from text prompts.
Because we can generate only those who make their Cameo avatars public, we collect only three subjects at the moment who are famous and easy to collect real videos on the Web.
We will enlarge our dataset so that it includes more subjects in the future. 

For each subject, we collect eight real videos of different scenes, then we split them into four videos each for reference and testing. 
For each video of the reference set, we manually split it into 10 eight-second clips, ensuring that each clip includes only the portions in which the subject is actually speaking and that the subject's face can be detected to extract facial landmarks~\cite{fan} on all the frames. 
Regarding the test set, we split it into three eight-second clips in the same manner as the reference set. 

Then, we generate fake videos with Sora2. 
To produce a high-quality dataset reducing biases towards video contents, we tried to generate videos whose contents are similar to the original videos.
To this end, we used GPT-5 to generate the caption of a frame of each video and used Whisper~\cite{whisper} to generate the transcription of each speech.
After that, we input the captions and transcriptions into Sora2 to generate videos corresponding to the original videos, specifying the subjects by the Cameo feature.
Note that we replace some sensitive words (\ie, someone's name) in transcriptions with an abstract word such as ``a man'' because Sora2 refuses such sensitive prompts.
The S2CFP dataset is available on our project page.

\begin{table}[t]
    \centering
    \begin{adjustbox}{width=0.6\linewidth}
    \begin{tabular}{c|c}
        \toprule
        Hyperparameter & Value\\
        \midrule
        Optimizer & Adan~\cite{adan}\\
        Learning Rate & 4e-4\\
        Diffusion Steps & 1000\\
        $\beta$ schedule & Linear \\
        Video Duration & 8 seconds \\
        Video FPS & 25 \\
        Expression Dimension & 53 \\
        Transformer Dimension & 512\\
        MLP Dimension &1024\\
        Num Heads & 8\\
        Num Layers & 8\\
        Dropout & 0.1\\
        Classifier-Free Dropout & 0.25\\
        Adapter Dimension & 512\\
        Adapter Length & 8\\
        \bottomrule
    \end{tabular}
    \end{adjustbox}
    \caption{\textbf{Hyperparameters for \methodshortname.}}
    \label{tb:hyperparam}
\end{table}

\section{Curated Dataset for Pre-training}
\label{sec:curated_dataset_sup}
We describe the details of curation of our pre-training dataset.
The overview of the dataset is shown in Fig.~\ref{fig:pretrain_statistics}. 
Note that we use Whisper~\cite{whisper} to detect language spoken in videos.
The specific data sources and pre-processing are as follows:

\noindent\textbf{VoxCeleb2~\cite{voxceleb2}.}
This dataset is one of the large-scale datasets of talking head videos from YouTube. 
The videos are already cropped into a small region around the face. 
To improve the quality of the samples from this dataset, we only adopt identity-consistent videos, ensuring that a single identity appears in each video. 
We compute the identity similarity by ArcFace~\cite{arcface} between all the pairs of frames and then exclude videos that include one or more frames with a lower identity similarity than 0.4.
As a result, we obtain $140,194$ videos. 

\noindent\textbf{AVSpeech~\cite{avspeech}.}
This dataset is also a video dataset of talking people from YouTube. 
We clip the talking parts using the provided annotations. 
Because the videos are provided without cropping, we track the faces by checking the overlap of the bounding boxes of the adjacent frames. 
As a result, we obtain $45,706$ videos. 

\noindent\textbf{Acappella~\cite{acappella}.}
This dataset provides videos of singing people from YouTube. 
Each video contains not only singing parts but also talking parts; we adopt both parts to enlarge our dataset. 
We track and crop the videos in the same manner as the pre-processing for the AVSpeech dataset. 
As a result, we obtain $14,459$ videos.

\section{Implementation Details}
\subsection{\methodname Model}
\label{sec:imple_ours_sup}
\noindent\textbf{Network Architecture.}
We adopt a similar network with DiT~\cite{dit} and EDGE~\cite{tseng2023edge}, but newly introduce TiLM layers for sequential conditioning described in Sec.~\ref{sec:model}.
The hyperparameters of our model are shown in Table~\ref{tb:hyperparam}.

\noindent\textbf{Condition Guidance.}
Similar to classifier-free guidance (CFG)~\cite{cfdg}, we train our model with learnable unconditional vectors for both audio and identity conditions to control the strength of conditions during inference.
Importantly, we empirically find that, in contrast to generative tasks (\eg, text-to-image synthesis~\cite{ldm}) where models aim to emphasize conditions, the large strength of guidance harms the detection performance.
We set the strengths $s_{a}$ and $s_{c}$ for audio and identity conditions to 0.5 and 0.25, respectively:
\begin{equation}
\label{eq:cfg_audio}
\epsilon_{\hat{\theta}_1}^{s_a}(\boldsymbol{z}_t^{1:L},t,\boldsymbol{w}^{1:H}) = \epsilon_{\hat{\theta}_1}(\boldsymbol{z}_t^{1:L},t) + s_{a}\delta_{a},
\end{equation}
\begin{equation}
\delta_{a} = \epsilon_{\hat{\theta}_1}(\boldsymbol{z}_t^{1:L},t,\boldsymbol{w}^{1:H}) - \epsilon_{\hat{\theta}_1}(\boldsymbol{z}_t^{1:L},t)
\end{equation}
\begin{equation}
\label{eq:cfg_identity}
\epsilon_{\hat{\theta}}^{\{s_a,s_c\}}(\boldsymbol{z}_t^{1:L},t,\boldsymbol{w}^{1:H},\boldsymbol{c}) = \epsilon_{\hat{\theta}_1}^{s_a}(\boldsymbol{z}_t^{1:L},t,\boldsymbol{w}^{1:H}) + s_{c}\delta_{c},
\end{equation}
\begin{equation}
\delta_{c} = \epsilon_{\hat{\theta}}(\boldsymbol{z}_t^{1:L},t,\boldsymbol{w}^{1:H},\boldsymbol{c}) - \epsilon_{\hat{\theta}_1}(\boldsymbol{z}_t^{1:L},t,\boldsymbol{w}^{1:H})
\end{equation}
We denote $\epsilon_{\hat{\theta}_1}^{s_a}(\boldsymbol{z}_t^{1:L},t,\boldsymbol{w}^{1:H})$ and $\epsilon_{\hat{\theta}}^{\{s_a,s_c\}}(\boldsymbol{z}_t^{1:L},t,\boldsymbol{w}^{1:H},\boldsymbol{c})$ as $\epsilon_{\hat{\theta}_1}(\boldsymbol{z}_t^{1:L},t,\boldsymbol{w}^{1:H})$ and $\epsilon_{\hat{\theta}}(\boldsymbol{z}_t^{1:L},t,\boldsymbol{w}^{1:H},\boldsymbol{c})$ in Eq.~\eqref{eq:authentication} in the main paper for simplicity, respectively.

\noindent\textbf{3DMM Extraction.}
For pre-training, we assign a single shape shared within each clip, assuming that different videos have different face shapes, although some identities overlap with each other. 
For personalization and authentication, we extract a single shape shared between videos considered to belong to an identical subject. 
This is achieved by using the shape extracted from the first reference video as a fixture in optimizations for all other videos.

\subsection{Baselines}
\label{sec:imple_previous_sup}

\noindent\textbf{EfficientNet-b4.}
We train the model on the FF++~\cite{ffpp} dataset, including real videos and their manipulated ones by Deepfakes~\cite{deepfake-faceswap}, Face2Face2~\cite{face2face}, FaceSwap~\cite{faceswap}, and NeuralTextures~\cite{neuraltextures} using the same pre-process, augmentations, and inference strategy as SBI~\cite{sbi}.

\noindent\textbf{Face X-ray.}
Because there is no official implementation, we re-implement it. 
We strictly follow the training setting of the original paper; we freeze the backbone~\cite{hrnet} for the first 50K iterations and then update all the layers for the remaining 150K iterations on real and blended images.
We generate the blended images using the author's unofficial implementation\footnote{\url{https://github.com/AlgoHunt/Face-Xray}}.
\noindent\textbf{UCF.}
We use the DeepfakeBench's implementation~\cite{dfbench}.

\noindent\textbf{Others.} We directly adopt their official implementations.

\begin{table}[t]
    \centering
    \begin{adjustbox}{width=1.0\linewidth}
    \begin{tabular}{lcccc|c} \toprule
      
      Setting& DF-TIMIT&DFDCP & KoDF & IDForge &Avg\\
      \midrule
        $t \sim \mathcal{U}[1, 1000]$ & \textbf{99.94} & 91.68 & 96.02 & 90.63 & 94.57\\
        $t \sim \mathcal{U}[101, 900]$& 99.84 & 92.84 & \textbf{96.14} & 91.96 & 95.20 \\
        \rowcolor[gray]{0.9}
        $t \sim \mathcal{U}[201, 800]$& 99.72 & 93.45 & 95.31 & 92.40 & \textbf{95.22} \\
        $t \sim \mathcal{U}[301, 700]$& 99.33 & \textbf{93.48} & 94.07 & \textbf{92.41} & 94.82\\
      \bottomrule
    \end{tabular}
    \end{adjustbox}
  \caption{\textbf{Study on timestep sampling.} Excluding early and late timesteps where identity information is not effective for denoising improves the results. Our default setting is highlighted in \colorbox[gray]{0.9}{gray}.}
  \label{tb:ablation_sampling}
\end{table}

\section{Additional Experiments}
\label{sec:additional_results}

\noindent\textbf{Timestep Sampling.}
We explore the sampling strategy of diffusion timesteps during authentication described in Sec.~\ref{sec:authentication}.
We evaluate variants of our method with different sets of timesteps, \ie, $[1, 1000]$, $[101, 900]$, $[201, 800]$, and $[301, 700]$ in Table~\ref{tb:ablation_sampling}.
It can be observed that excluding both sides of timesteps from authentication helps our model detect deepfakes more accurately (94.57\% by $[1, 1000]$ vs.~95.22\% by $[201, 800]$ in the average AUC). 
However, the excessive exclusion harms the detection performance (95.22\% by $[201, 800]$ vs. 95.20\% by $[301, 700]$ in the average AUC). 
We recommend using the range $[201, 800]$ by default for stable performance on different datasets.

\begin{table}[t]
    \centering
    \begin{minipage}{0.38\linewidth}
    \begin{adjustbox}{width=1.0\linewidth}
    \centering
    \begin{tabular}{lcc}
        \toprule
        Method & Threshold & KoDF\\ 
        \midrule
        AltFreezing &0.5&86.75\\
        DFD-FCG & 0.5 & 86.57\\ 
        \rowcolor[gray]{0.9}
        Ours & $\mu + \sigma$ & 89.73 \\
        \rowcolor[gray]{0.9}
        Ours & $\mu + 2\sigma$ & \textbf{90.85}\\
        \rowcolor[gray]{0.9}
        Ours & $\mu + 3\sigma$ & 86.38\\
        \bottomrule
    \end{tabular}
    \end{adjustbox}
    \caption{\textbf{ACC on KoDF.}}
    \label{tb:accuracy}
    \end{minipage}
\hfill
\begin{minipage}{0.59\linewidth}
    \centering
    \begin{adjustbox}{width=1.0\linewidth}
    \begin{tabular}{lccc}
        \toprule
        Method & \#Params & Time[s] & Avg\\ 
        \midrule
        LipForensics &36M&0.67&88.92\\
        AltFreezing &27M&3.56&90.34\\
        SBI  &18M&0.82&85.50\\
        ForensicsAdapter &435M&0.37&90.33\\
        \rowcolor[gray]{0.9}
        Ours & 31M + 36M &22.2 + 23.6 & \textbf{95.22}\\
        \bottomrule
    \end{tabular}
    \end{adjustbox}
    \caption{\textbf{Model complexity analysis.}}
    \label{tb:complexity}
    \end{minipage}
\end{table}

\noindent\textbf{Thresholding.}
In the main paper, we focus on the potential discriminative ability by evaluating models with a thresholding-free metric, \ie, AUC.
Here, we describe how to decide whether videos are real or fake.
We assume that the prediction scores from each subject follow a subject-specific Gaussian distribution.
Therefore, we compute the mean $\mu$ and the unbiased standard deviation $\sigma$ from a validation set including eight real videos of the subject.
Then, we set a threshold to divide the real and fake classes.
To evaluate the effectiveness, we compute the accuracy (ACC) on KoDF in Table~\ref{tb:accuracy} with thresholds $\mu + \sigma$, $\mu + 2\sigma$, and $\mu + 3\sigma$.
We also show the results of AltFreezing and DFD-FCG, which perform best on KoDF as shown in Table~\ref{tb:cross_dataset}, with the threshold simply set to $0.5$.
Our method achieves consistent accuracy with diverse thresholds, which indicates that our model accurately distinguishes real videos from deepfakes.
Note that it is difficult to perform this experiment on the DF-TIMIT and DFDCP datasets because they have a limited number of videos for each subject to prepare validation sets.

\noindent\textbf{Complexity Analysis.}
We compare our model with the state-of-the-art methods in terms of the model complexity in Table~\ref{tb:complexity}.
We compute the number of parameters and the inference time, excluding data loading, per video with eight seconds on a single NVIDIA A100 GPU.
Note that we follow the official inference strategy for AltFreezing, which can take more time than straightforward inference.
Our model has 47M parameters consisting of 31M of SPECTRE and 36M of our diffusion model, which is much smaller than ForensicsAdapter.
Our inference takes 22.2 and 23.6 seconds for 3DMM extraction and diffusion authentication, respectively, which is slower than the previous methods. 
We do not focus on the optimization of inference time in this paper, and it could be improved as follows in future: 
First, because our 3DMM extraction strategy performs iterative refinement, taking a long time, developing a feed-forward extraction model that directly predicts disentangled FLAME parameters can drastically reduce the overhead.
Second, we can reduce the diffusion costs by using a smaller number of noise sequences; we observe in Fig.~\ref{fig:comprehensive_analysis}(b) that using a quarter of our default number of noise sequences achieves the same AUC on DFDCP.

\begin{table}[t]
    \centering
    \begin{adjustbox}{width=1.0\linewidth}
    \begin{tabular}{lcccc|c} \toprule
      
      Setting& DF-TIMIT&DFDCP & KoDF & IDForge &Avg\\
      \midrule
        w/o Audio & 99.46 & 90.76 &93.41 &91.24 & 93.72\\
        Ours& \textbf{99.72} & \textbf{93.45} & \textbf{95.31} & \textbf{92.40} & \textbf{95.22}\\
      \bottomrule
    \end{tabular}
    \end{adjustbox}
  \caption{\textbf{Effect of audio conditioning.}}
  \label{tb:ablation_audio}
\end{table}

\noindent\textbf{Effect of Audio Conditioning.}
We demonstrate in our framework that conditioning reconstruction on audio helps detection accuracy.
To this end, we perform inference by replacing audio conditions with the learned unconditional vector (see App.~\ref{sec:imple_ours_sup} and CFG~\cite{cfdg} for the unconditional vector).
As shown in Table~\ref{tb:ablation_audio}, our model without audio conditions drops the AUCs on all the test sets.
This result indicates that audio conditioning helps the prediction of the expression coefficients and thus improves detection performance. 
Notably, our method without audio \textbf{still achieves the state-of-the-art generalization ability} in Table~\ref{tb:cross_dataset}.

\noindent\textbf{Comprehensive Comparison on S2CFP.}
We show the result on S2CFP with all the baselines from Table~\ref{tb:cross_dataset} in Table~\ref{tb:sora_all}.
In addition, we refer to the state-of-the-art diffusion-generated image detectors~\cite{dire,aeroblade,bfree}.
DIRE~\cite{dire} argues that diffusion models reconstruct diffusion-generated images more precisely than real images, which enables general diffusion-generated image detection. 
AEROBLADE~\cite{aeroblade} applies VAE reconstruction to expose latent-diffusion-generated images.
B-Free~\cite{bfree} carefully curates its training set so that there is no content difference between real and fake classes to mitigate biases towards image contents.
Even compared with these methods, specialized in diffusion-generated image detection, our method achieves the best result with a large margin.

\begin{table}[t]
    \centering
    \begin{adjustbox}{width=1.0\linewidth}
    \begin{tabular}{lccc|c} \toprule
    \multirow{2}{*}{Method}& \multicolumn{4}{c}{Test Set AUC (\%) on Each Subject}\\ 
        \cmidrule(lr){2-5}
      & @ijustine &@mcuban & @sama & \textbf{Avg}\\
      \midrule
        EfficientNet-b4&65.97 & 41.67 & 35.42 & 47.69 \\
        LipForensics & 48.61 & 56.94 & 33.33 & 46.29\\
        FTCN & 34.03 & 56.94 & 29.86 & 40.28\\
        RECCE & 46.53 & 66.67 & 49.31 & 54.17 \\
        RealForensics & 45.83 & 58.33 & 51.39 & 51.85\\
        AltFreezing & 27.78 & 38.19 & 15.97 & 27.31\\
        UCF & 36.11 &12.50 & 28.47 & 25.69\\
        LipFD &17.36 & 51.39 & 50.69 & 39.81\\
        FSFM & 40.97 & 76.39 & 70.14 & 62.50\\
        DFD-FCG & 36.11 & 53.47 & 56.25 & 48.61\\
        EFFORT & \underline{85.42} & 56.94 & 61.03 & 67.80\\
        Face X-ray &4.86 & 45.83 & 18.06 & 22.92\\
        SBI & 45.14 &45.83 & 50.69 & 47.22\\
        ICT & 71.53 & 78.47 & 34.72 & 61.57\\
        LAA-Net w/ SBI & 22.22 & 37.50 & 70.83 & 43.52\\
        ForensicsAdapter & 38.89 & 82.64 & 62.50 & 61.34\\
        ID-Reveal & 83.33 & \textbf{87.50} & 75.69 & \underline{81.02}\\
        AVAD & 9.03 & 00.00 & 15.97 & 8.33\\
        POI-Forensics & 41.67 & 43.75 & 72.22 & 52.55\\
        SpeechForensics & 54.86 & 64.58 & 63.89 & 61.11\\
        DIRE & 11.11 & 56.25 & 43.75 & 37.04\\
        AEROBLADE & 43.75 & 52.08 & \underline{91.67} & 62.50\\
        B-Free & 65.97 & 71.53 & 76.39 & 71.30 \\
        \rowcolor[gray]{0.9}
        Ours & \textbf{98.61} & \underline{84.72} & \textbf{100.00} & \textbf{94.44}\\
      \bottomrule
    \end{tabular}
    \end{adjustbox}
  \caption{\textbf{Comprehensive comparison on S2CFP.}}
  \label{tb:sora_all}
\end{table}

\begin{figure*}[t]
  \centering
  \begin{minipage}{1.0\linewidth}
  \centering
  \includegraphics[width=1.0\linewidth]{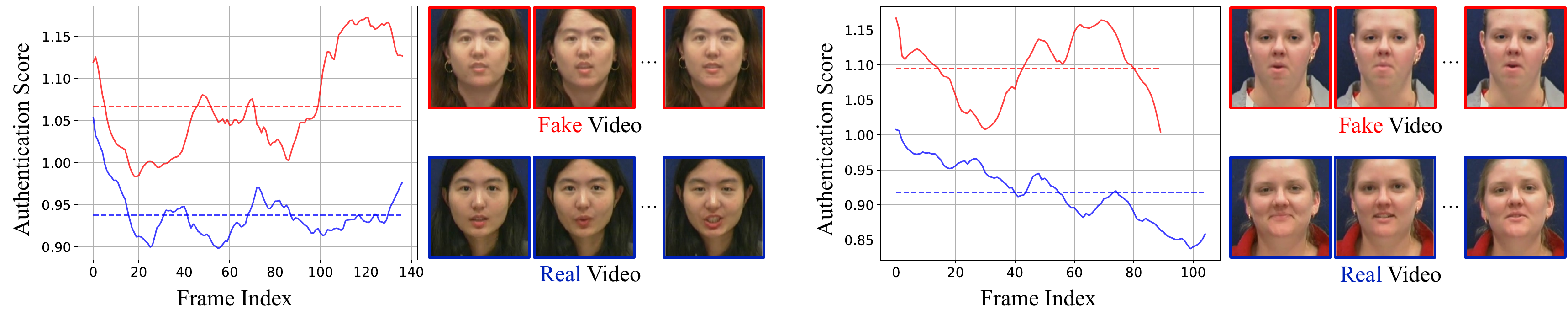}
  \par\vspace{-0.5em}
  \subcaption{On DF-TIMIT}
  \end{minipage}
  \begin{minipage}{1.0\linewidth}
  \centering
  \includegraphics[width=1.0\linewidth]{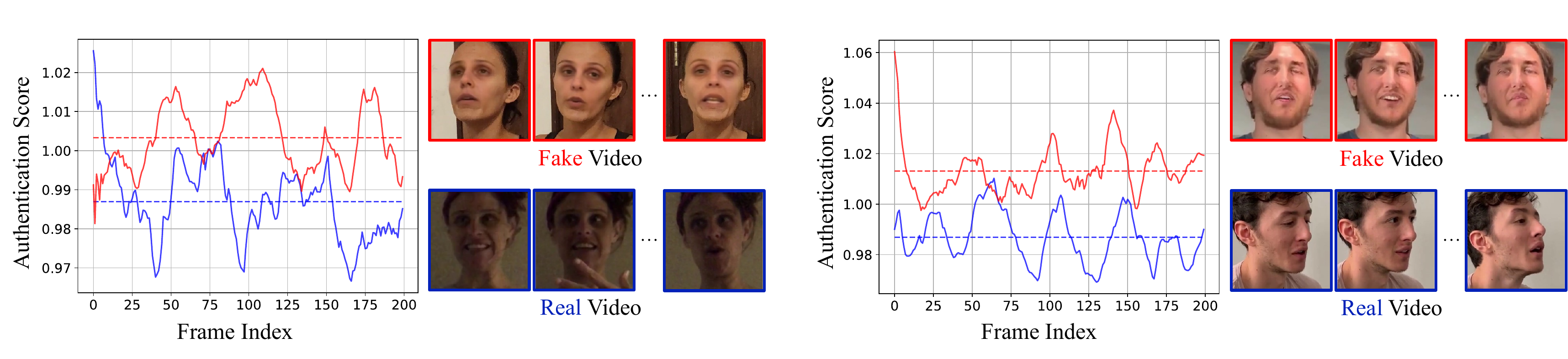}
  \par\vspace{-0.5em}
  \subcaption{On DFDCP}
  \end{minipage}
  \par\vspace{1em}
  \begin{minipage}{1.0\linewidth}
  \centering
  \includegraphics[width=1.0\linewidth]{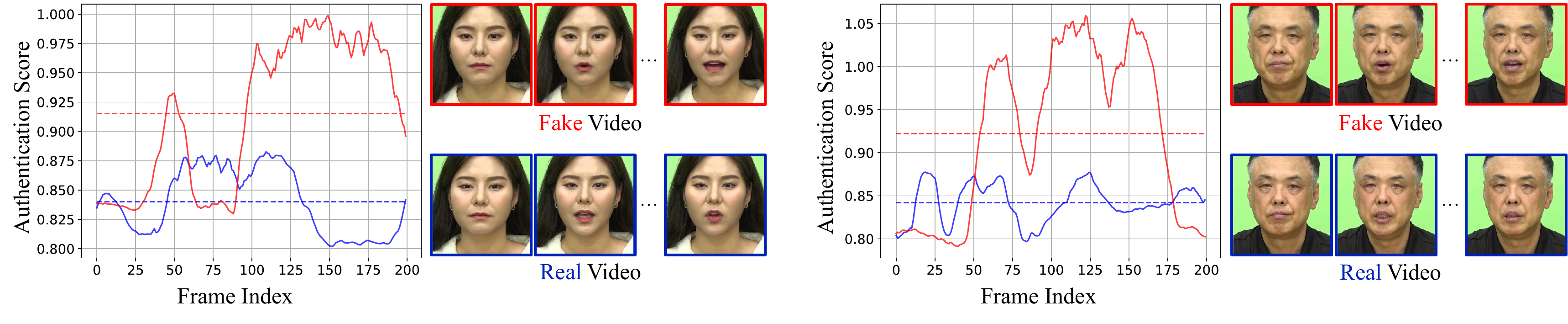}
  \par\vspace{-0.5em}
  \subcaption{On KoDF}
  \end{minipage}
  \par\vspace{1em}
  \begin{minipage}{1.0\linewidth}
  \centering
  \includegraphics[width=1.0\linewidth]{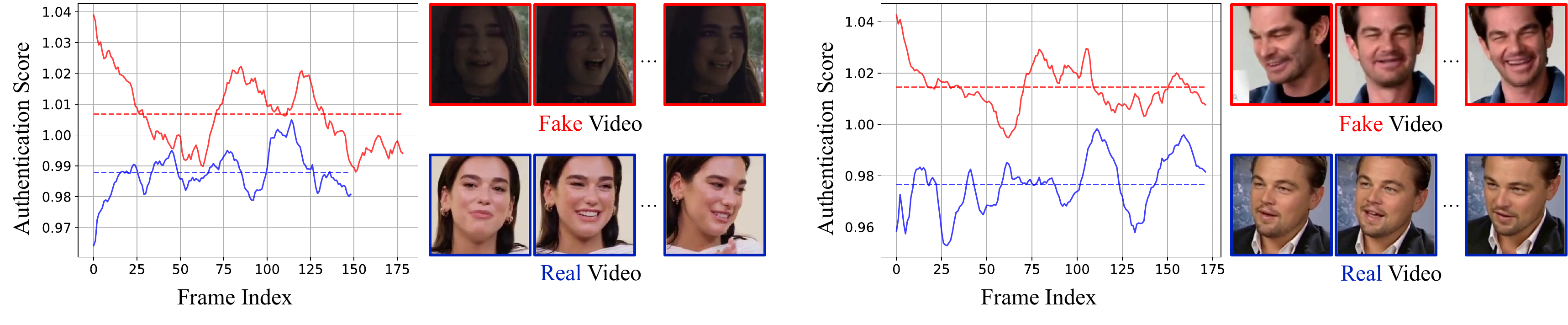}
  \par\vspace{-0.5em}
  \subcaption{On IDForge}
  \end{minipage}
  \par\vspace{1em}
  \begin{minipage}{1.0\linewidth}
  \centering
  \includegraphics[width=1.0\linewidth]{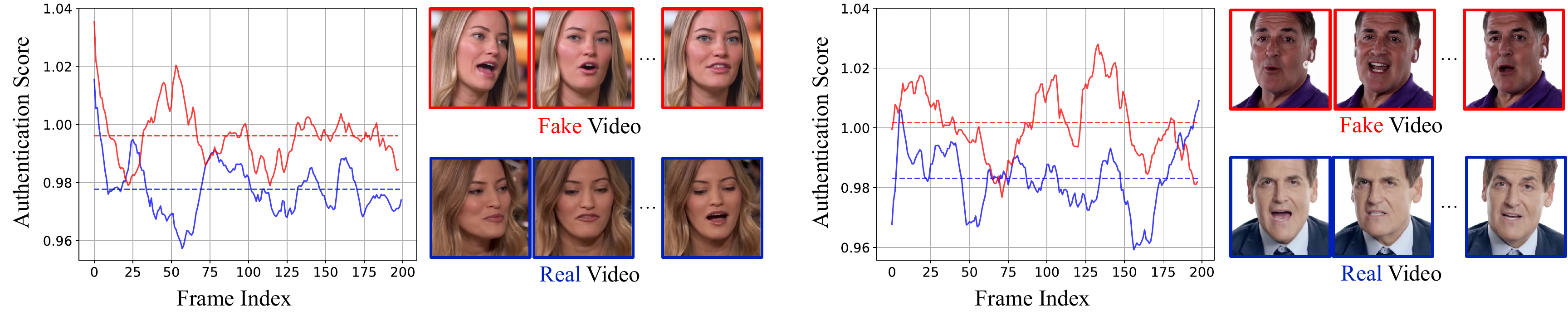}
  \par\vspace{-0.5em}
  \subcaption{On S2CFP}
  \end{minipage}
  \caption{\textbf{Additional visualizations.} The solid blue and red lines represent the authentication scores over the frame index of real videos and those of deepfakes mimicking the subjects, respectively; the dotted lines denote the corresponding averages.} 
  \label{fig:more_visualization}
\end{figure*}

\noindent\textbf{Additional Visualizations.}
We show additional examples of temporal authentication scores in Fig.~\ref{fig:more_visualization}.
Overall, our model performs well on a wide range of subjects and scenes and is robust to a variety of manipulations.
We obtain some important observations from the figures:
1) The authentication scores are slightly unstable at silent frames \eg, the beginning and ending of videos. 
2) Our model distinguishes real samples from fake ones even in cases where deepfakes' appearances are much closer to real ones, as seen on KoDF and S2CFP, which indicates that our method focuses on highly semantic talking identities for face forgery detection.

\clearpage
{
    \small
    \bibliographystyle{ieeenat_fullname}
    \bibliography{main}

@String(CVPR= {IEEE Conf. Comput. Vis. Pattern Recog.})

@String(ICCV= {Int. Conf. Comput. Vis.})

@String(ECCV= {Eur. Conf. Comput. Vis.})

@String(BMVC= {Brit. Mach. Vis. Conf.})

@String(TOG= {ACM Trans. Graph.})

@String(ICLR = {Int. Conf. Learn. Represent.})

@String(AAAI = {AAAI})

@String(CVPRW= {IEEE Conf. Comput. Vis. Pattern Recog. Worksh.})

@String(CVPR  = {CVPR})

@String(ICCV  = {ICCV})

@String(ECCV  = {ECCV})

@String(BMVC  =	{BMVC})

@String(TOG   = {ACM TOG})

@String(ICLR  = {ICLR})

@String(CVPRW= {CVPRW})

@InProceedings{facexray,
author = {Li, Lingzhi and Bao, Jianmin and Zhang, Ting and Yang, Hao and Chen, Dong and Wen, Fang and Guo, Baining},
title = {Face X-Ray for More General Face Forgery Detection},
booktitle = {CVPR},
year = {2020}
}

@inproceedings{pcl,
  title={Learning Self-Consistency for Deepfake Detection},
  author={Zhao, Tianchen and Xu, Xiang and Xu, Mingze and Ding, Hui and Xiong, Yuanjun and Xia, Wei},
  booktitle={ICCV},
  year={2021}
}

@inproceedings{lipforensics,
  title={Lips Don't Lie: A Generalisable and Robust Approach To Face Forgery Detection},
  author={Haliassos, Alexandros and Vougioukas, Konstantinos and Petridis, Stavros and Pantic, Maja},
  booktitle={CVPR},
  year={2021}
}

@inproceedings{idreveal,
  title={Id-Reveal: Identity-aware deepfake video detection},
  author={Cozzolino, Davide and R{\"o}ssler, Andreas and Thies, Justus and Nie{\ss}ner, Matthias and Verdoliva, Luisa},
  booktitle={ICCV},
  year={2021}
}

@InProceedings{poiforensics,
    author    = {Cozzolino, Davide and Pianese, Alessandro and Nie{\ss}ner, Matthias and Verdoliva, Luisa},
    title     = {Audio-Visual Person-of-Interest DeepFake Detection},
    booktitle = {CVPR Workshop},
    year      = {2023},
}

@inproceedings{agarwal2019protecting,
  title={Protecting world leaders against deep fakes.},
  author={Agarwal, Shruti and Farid, Hany and Gu, Yuming and He, Mingming and Nagano, Koki and Li, Hao},
  booktitle={CVPR Workshop},
  year={2019}
}

@inproceedings{agarwal2023watch,
  title={Watch those words: Video falsification detection using word-conditioned facial motion},
  author={Agarwal, Shruti and Hu, Liwen and Ng, Evonne and Darrell, Trevor and Li, Hao and Rohrbach, Anna},
  booktitle={WACV},
  year={2023}
}

@INPROCEEDINGS{agarwal2020detecting,
  author={Agarwal, Shruti and Farid, Hany and El-Gaaly, Tarek and Lim, Ser-Nam},
  booktitle={WIFS}, 
  title={Detecting Deep-Fake Videos from Appearance and Behavior}, 
  year={2020}}

@ARTICLE{fang2024stidnet,
  author={Fang, Mingqi and Yu, Lingyun and Xie, Hongtao and Tan, Qingfeng and Tan, Zhiyuan and Hussain, Amir and Wang, Zezheng and Li, Jiahong and Tian, Zhihong},
  journal={IEEE Transactions on Computational Social Systems}, 
  title={STIDNet: Identity-Aware Face Forgery Detection With Spatiotemporal Knowledge Distillation}, 
  year={2024}}

@inproceedings{sbi,
  title={Detecting Deepfakes with Self-Blended Images},
  author={Shiohara, Kaede and Yamasaki, Toshihiko},
  booktitle={CVPR},
  year={2022}
}

@inproceedings{mtsbi,
    author = {Huang, Po-Han and Han, Yue-Hua and Chu, Ernie and Chen, Jun-Cheng and Hua, Kai-Lung},
    title = {Multi-Task Self-Blended Images for Face Forgery Detection},
    year = {2024},
    booktitle = {MM Asia}
}

@InProceedings{fwa,
    author = {Li, Yuezun and Lyu, Siwei},
    title = {Exposing DeepFake Videos By Detecting Face Warping Artifacts},
    booktitle = {CVPR Workshop},
    year = {2019}
}

@inproceedings{ram,
    author = {Tian, Jiahe and Yu, Cai and Wang, Xi and Chen, Peng and Xiao, Zihao and Dai, Jiao and Han, Jizhong and Chai, Yesheng},
    title = {Real Appearance Modeling for More General Deepfake Detection},
    year = {2024},
    booktitle = {ECCV}
}

@InProceedings{recce,
    author    = {Cao, Junyi and Ma, Chao and Yao, Taiping and Chen, Shen and Ding, Shouhong and Yang, Xiaokang},
    title     = {End-to-End Reconstruction-Classification Learning for Face Forgery Detection},
    booktitle = {CVPR},
    year      = {2022}
}

@INPROCEEDINGS{dire,
author = { Wang, Zhendong and Bao, Jianmin and Zhou, Wengang and Wang, Weilun and Hu, Hezhen and Chen, Hong and Li, Houqiang },
booktitle = {ICCV},
title = {DIRE for Diffusion-Generated Image Detection },
year = {2023}}

@inproceedings{aeroblade,
  title={Aeroblade: Training-free detection of latent diffusion images using autoencoder reconstruction error},
  author={Ricker, Jonas and Lukovnikov, Denis and Fischer, Asja},
  booktitle={CVPR},
  year={2024}
}

@InProceedings{bfree,
    author    = {Guillaro, Fabrizio and Zingarini, Giada and Usman, Ben and Sud, Avneesh and Cozzolino, Davide and Verdoliva, Luisa},
    title     = {A Bias-Free Training Paradigm for More General AI-generated Image Detection},
    booktitle = {CVPR},
    year      = {2025},
}

@inproceedings{sladd,
    author = {Liang Chen and Yong Zhang and Yibing Song and Lingqiao Liu and Jue Wang},
    title = {Self-supervised Learning of Adversarial Examples: Towards Good Generalizations for DeepFake Detections}, 
    booktitle = {CVPR},  
    year = {2022}
}

@inproceedings{realforensics,
  title={Leveraging Real Talking Faces via Self-Supervision for Robust Forgery Detection},
  author={Haliassos, Alexandros and Mira, Rodrigo and Petridis, Stavros and Pantic, Maja},
  booktitle={CVPR},
  year={2022}
}

@inproceedings{ucf,
 title={Ucf: Uncovering common features for generalizable deepfake detection},
 author={Yan, Zhiyuan and Zhang, Yong and Fan, Yanbo and Wu, Baoyuan},
 booktitle={ICCV},
 year={2023}
}

@InProceedings{laanet,
    author    = {Nguyen, Dat and Mejri, Nesryne and Singh, Inder Pal and Kuleshova, Polina and Astrid, Marcella and Kacem, Anis and Ghorbel, Enjie and Aouada, Djamila},
    title     = {LAA-Net: Localized Artifact Attention Network for Quality-Agnostic and Generalizable Deepfake Detection},
    booktitle = {CVPR},
    year      = {2024}
}

@InProceedings{altfreezing,
    author    = {Wang, Zhendong and Bao, Jianmin and Zhou, Wengang and Wang, Weilun and Li, Houqiang},
    title     = {AltFreezing for More General Video Face Forgery Detection},
    booktitle = {CVPR},
    year      = {2023}
}

@inproceedings{ftcn,
  title={Exploring Temporal Coherence for More General Video Face Forgery Detection},
  author={Zheng, Yinglin and Bao, Jianmin and Chen, Dong and Zeng, Ming and Wen, Fang},
  booktitle={ICCV},
  year={2021}
}

@inproceedings{dfdfcg,
      title={Towards More General Video-based Deepfake Detection through Facial Component Guided Adaptation for Foundation Model},
      author={Han, Yue-Hua and Huang, Tai-Ming and Hua, Kai-Lung and Chen, Jun-Cheng},
      booktitle={CVPR},
      year={2025}
}

@InProceedings{forensicsadapter,
    author    = {Cui, Xinjie and Li, Yuezun and Luo, Ao and Zhou, Jiaran and Dong, Junyu},
    title     = {Forensics Adapter: Adapting CLIP for Generalizable Face Forgery Detection},
    booktitle = {CVPR},
    year      = {2025}
}

@inproceedings{fsfm,
  title={Fsfm: A generalizable face security foundation model via self-supervised facial representation learning},
  author={Wang, Gaojian and Lin, Feng and Wu, Tong and Liu, Zhenguang and Ba, Zhongjie and Ren, Kui},
  booktitle={CVPR},
  year={2025}
}

@inproceedings{
    effort,
    title={Orthogonal Subspace Decomposition for Generalizable {AI}-Generated Image Detection},
    author={Zhiyuan Yan and Jiangming Wang and Peng Jin and Ke-Yue Zhang and Chengchun Liu and Shen Chen and Taiping Yao and Shouhong Ding and Baoyuan Wu and Li Yuan},
    booktitle={ICML},
    year={2025},
}

@INPROCEEDINGS {joint,
    author = {Zhou, Yipin and Lim, Ser-Nam},
    booktitle = {ICCV},
    title = {Joint Audio-Visual Deepfake Detection},
    year = {2021},
}

@InProceedings{avff,
    author    = {Oorloff, Trevine and Koppisetti, Surya and Bonettini, Nicol\`o and Solanki, Divyaraj and Colman, Ben and Yacoob, Yaser and Shahriyari, Ali and Bharaj, Gaurav},
    title     = {AVFF: Audio-Visual Feature Fusion for Video Deepfake Detection},
    booktitle = {CVPR},
    year      = {2024},
}

@inproceedings{avlgi,
    author = {Wang, Yifan and Wu, Xuecheng and Zhang, Jia and Jing, Mohan and Lu, Keda and Yu, Jun and Su, Wen and Gao, Fang and Liu, Qingsong and Sun, Jianqing and Liang, Jiaen},
    title = {Building Robust Video-Level Deepfake Detection via Audio-Visual Local-Global Interactions},
    year = {2024},
    booktitle = {MM},
}

@inproceedings{liu2024lips,
 author = {Liu, Weifeng and She, Tianyi and Liu, Jiawei and Li, Boheng and Yao, Dongyu and Liang, Ziyou and Wang, Run},
 booktitle = {NeurIPS},
 title = {Lips Are Lying: Spotting the Temporal Inconsistency between Audio and Visual in Lip-Syncing DeepFakes},
 year = {2024}
}

@inproceedings{avad,
  title={Self-supervised video forensics by audio-visual anomaly detection},
  author={Feng, Chao and Chen, Ziyang and Owens, Andrew},
  booktitle={CVPR},
  year={2023}
}

@article{factor,
  title={Detecting Deepfakes Without Seeing Any},
  author={Reiss, Tal and Cavia, Bar and Hoshen, Yedid},
  journal={arXiv:2311.01458},
  year={2023}
}

@INPROCEEDINGS{ocfd,
  author={Khalid, Hasam and Woo, Simon S.},
  booktitle={CVPRW}, 
  title={OC-FakeDect: Classifying Deepfakes Using One-class Variational Autoencoder}, 
  year={2020}
}

@inproceedings{
    speechforensics,
    title={SpeechForensics: Audio-Visual Speech Representation Learning for Face Forgery Detection},
    author={Yachao Liang and Min Yu and Gang Li and Jianguo Jiang and Boquan Li and Feng Yu and Ning Zhang and Xiang Meng and Weiqing Huang},
    booktitle={NeurIPS},
    year={2024}
}

@article{avlipsyncplus,
  title={AV-Lip-Sync+: Leveraging AV-HuBERT to exploit multimodal inconsistency for video deepfake detection},
  author={Shahzad, Sahibzada Adil and Hashmi, Ammarah and Peng, Yan-Tsung and Tsao, Yu and Wang, Hsin-Min},
  journal={arXiv:2311.02733},
  year={2023}
}

@inproceedings{ict,
  title={Protecting Celebrities from DeepFake with Identity Consistency Transformer},
  author={Dong, Xiaoyi and Bao, Jianmin and Chen, Dongdong and Zhang, Ting and Zhang, Weiming and Yu, Nenghai and Chen, Dong and Wen, Fang and Guo, Baining},
  booktitle={CVPR},
  year={2022}
}

@inproceedings{gan,
  title={Generative adversarial nets},
  author={Goodfellow, Ian and Pouget-Abadie, Jean and Mirza, Mehdi and Xu, Bing and Warde-Farley, David and Ozair, Sherjil and Courville, Aaron and Bengio, Yoshua},
  booktitle={NeurIPS},
  year={2014}
}

@article{dcgan,
  title={Unsupervised representation learning with deep convolutional generative adversarial networks},
  author={Radford, Alec and Metz, Luke and Chintala, Soumith},
  journal={arXiv:1511.06434},
  year={2015}
}

@inproceedings{stylegan,
  title={A style-based generator architecture for generative adversarial networks},
  author={Karras, Tero and Laine, Samuli and Aila, Timo},
  booktitle={CVPR},
  year={2019}
}

@inproceedings{efficientnet,
  title={Efficientnet: Rethinking model scaling for convolutional neural networks},
  author={Tan, Mingxing and Le, Quoc},
  booktitle={ICML},
  year={2019}
}

@article{hrnet,
  title={Deep high-resolution representation learning for visual recognition},
  author={Wang, Jingdong and Sun, Ke and Cheng, Tianheng and Jiang, Borui and Deng, Chaorui and Zhao, Yang and Liu, Dong and Mu, Yadong and Tan, Mingkui and Wang, Xinggang and others},
  journal={TPAMI},
  year={2020}
}

@article{vae,
  title={Auto-encoding variational bayes},
  author={Kingma, Diederik P and Welling, Max},
  journal={arXiv:1312.6114},
  year={2013}
}

@inproceedings{vqvae,
  title={Neural discrete representation learning},
  author={Oord, Aaron van den and Vinyals, Oriol and Kavukcuoglu, Koray},
  booktitle={NeurIPS},
  year={2017}
}

@inproceedings{ldm,
      title={High-Resolution Image Synthesis with Latent Diffusion Models}, 
      author={Robin Rombach and Andreas Blattmann and Dominik Lorenz and Patrick Esser and Björn Ommer},
      year={2022},
    booktitle = {CVPR}
}

@inproceedings{ddpm,
  title={Denoising diffusion probabilistic models},
  author={Ho, Jonathan and Jain, Ajay and Abbeel, Pieter},
  booktitle={NeurIPS},
  year={2020}
}

@inproceedings{ddim,
    title={Denoising Diffusion Implicit Models},
    author={Jiaming Song and Chenlin Meng and Stefano Ermon},
    booktitle={ICLR},
    year={2021}
}

@inproceedings{kdiffusion,
    author = {Karras, Tero and Aittala, Miika and Aila, Timo and Laine, Samuli},
    booktitle = {NeurIPS},
    title = {Elucidating the Design Space of Diffusion-Based Generative Models},
    year = {2022}
}

@inproceedings{imagen,
    author = {Saharia, Chitwan and Chan, William and Saxena, Saurabh and Li, Lala and Whang, Jay and Denton, Emily L and Ghasemipour, Kamyar and Gontijo Lopes, Raphael and Karagol Ayan, Burcu and Salimans, Tim and Ho, Jonathan and Fleet, David J and Norouzi, Mohammad},
    booktitle = {NeurIPS},
    title = {Photorealistic Text-to-Image Diffusion Models with Deep Language Understanding},
    year = {2022}
}

@inproceedings{wav2vec, 
author = {Baevski, Alexei and Zhou, Henry and Mohamed, Abdelrahman and Auli, Michael}, 
title = {wav2vec 2.0: a framework for self-supervised learning of speech representations}, 
year = {2020}, 
booktitle = {NeurIPS} 
}

@article{adan,
  title={Adan: Adaptive nesterov momentum algorithm for faster optimizing deep models},
  author={Xie, Xingyu and Zhou, Pan and Li, Huan and Lin, Zhouchen and Yan, Shuicheng},
  journal={TPAMI},
  year={2024},
}

@inproceedings{dit,
  title={Scalable Diffusion Models with Transformers},
  author={William Peebles and Saining Xie},
  booktitle={ICCV},
  year={2023}
}

@inproceedings{film,
  title={Film: Visual reasoning with a general conditioning layer},
  author={Perez, Ethan and Strub, Florian and De Vries, Harm and Dumoulin, Vincent and Courville, Aaron},
  booktitle={AAAI},
  year={2018}
}

@InProceedings{arcface,
author = {Deng, Jiankang and Guo, Jia and Xue, Niannan and Zafeiriou, Stefanos},
title = {ArcFace: Additive Angular Margin Loss for Deep Face Recognition},
booktitle = {CVPR},
year = {2019}
}

@InProceedings{hyperiqa,
author = {Su, Shaolin and Yan, Qingsen and Zhu, Yu and Zhang, Cheng and Ge, Xin and Sun, Jinqiu and Zhang, Yanning},
title = {Blindly Assess Image Quality in the Wild Guided by a Self-Adaptive Hyper Network},
booktitle = {CVPR},
year = {2020}
}

@inproceedings{attention,
  title={Attention is all you need},
  author={Vaswani, Ashish and Shazeer, Noam and Parmar, Niki and Uszkoreit, Jakob and Jones, Llion and Gomez, Aidan N and Kaiser, {\L}ukasz and Polosukhin, Illia},
  booktitle={NeurIPS},
  year={2017}
}

@inproceedings{llamaadapter,
    title={{LL}a{MA}-Adapter: Efficient Fine-tuning of Large Language Models with Zero-initialized Attention},
    author={Renrui Zhang and Jiaming Han and Chris Liu and Aojun Zhou and Pan Lu and Yu Qiao and Hongsheng Li and Peng Gao},
    booktitle={ICLR},
    year={2024},
}

@inproceedings{adam,
  title={Adam: A Method for Stochastic Optimization},
  author={Kingma, Diederik P and Ba, Jimmy},
  booktitle={ICLR},
  year={2015}
}

@article{layernorm,
    title={Layer Normalization}, 
    author={Jimmy Lei Ba and Jamie Ryan Kiros and Geoffrey E. Hinton},
    year={2016},
    journal={arXiv:1607.06450}
}

@inproceedings{whisper,
  title={Robust speech recognition via large-scale weak supervision},
  author={Radford, Alec and Kim, Jong Wook and Xu, Tao and Brockman, Greg and McLeavey, Christine and Sutskever, Ilya},
  booktitle={International conference on machine learning},
  year={2023},
  organization={PMLR}
}

@inproceedings{cfdg,
  title={Classifier-Free Diffusion Guidance},
  author={Ho, Jonathan and Salimans, Tim},
  booktitle={NeurIPS Workshop},
  year={2021}
}

@inproceedings{tseng2023edge,
  title={Edge: Editable dance generation from music},
  author={Tseng, Jonathan and Castellon, Rodrigo and Liu, Karen},
  booktitle={CVPR},
  year={2023}
}

@InProceedings{diffusionclassifier,
    author    = {Li, Alexander C. and Prabhudesai, Mihir and Duggal, Shivam and Brown, Ellis and Pathak, Deepak},
    title     = {Your Diffusion Model is Secretly a Zero-Shot Classifier},
    booktitle = {ICCV},
    year      = {2023}
}

@inproceedings{chen2024robust,
  title={Robust classification via a single diffusion model},
  author={Chen, Huanran and Dong, Yinpeng and Wang, Zhengyi and Yang, Xiao and Duan, Chengqi and Su, Hang and Zhu, Jun},
  booktitle={ICML},
  year={2024}
}

@Misc{deepfake-faceswap,
  howpublished = "\url{https://github.com/deepfakes/faceswap}",
  note={Accessed: 2025-11-4},
  key = {Deepfakes}
}

@Misc{faceswap,
  key        = "FaceSwap",
  howpublished = "\url{https://github.com/MarekKowalski/FaceSwap/}",
  note={Accessed: 2025-11-4}
}

@inproceedings{face2face,
  author={Justus {Thies} and Michael {Zollh\"{o}fer} and Marc {Stamminger} and Christian {Theobalt} and Matthias {Nie{\ss}ner}},
  booktitle={CVPR}, 
  title={Face2Face: Real-Time Face Capture and Reenactment of RGB Videos}, 
  year={2016}}

@article{neuraltextures,
  author = {Thies, Justus and Zollh{\"o}fer, Michael and Nie{\ss}ner, Matthias},
  title = {Deferred Neural Rendering: Image Synthesis using Neural Textures},
  journal={TOG},
  year={2019}
}

@InProceedings{blendface,
    author    = {Shiohara, Kaede and Yang, Xingchao and Taketomi, Takafumi},
    title     = {BlendFace: Re-designing Identity Encoders for Face-Swapping},
    booktitle = {ICCV},
    year      = {2023}
}

@inproceedings{simswap,
  author    = {Renwang Chen and
               Xuanhong Chen and
               Bingbing Ni and
               Yanhao Ge},
  title     = {{SimSwap: An Efficient Framework For High Fidelity Face Swapping}},
  booktitle = {MM},
  year      = {2020}
}

@InProceedings{faceshifter,
author = {Li, Lingzhi and Bao, Jianmin and Yang, Hao and Chen, Dong and Wen, Fang},
title = {Advancing High Fidelity Identity Swapping for Forgery Detection},
booktitle = {CVPR},
year = {2020}
}

@article{dfl,
    author = {Liu, Kunlin and Perov, Ivan and Gao, Daiheng and Chervoniy, Nikolay and Zhou, Wenbo and Zhang, Weiming},
    title = {Deepfacelab: Integrated, flexible and extensible face-swapping framework},
    year = {2023},
    journal = {Pattern Recogn.},
}

@inproceedings{wav2lip,
  title={A lip sync expert is all you need for speech to lip generation in the wild},
  author={Prajwal, KR and Mukhopadhyay, Rudrabha and Namboodiri, Vinay P and Jawahar, CV},
  booktitle={MM},
  year={2020}
}

@article{audiodriven,
  title     = {Audio-driven talking face video generation with learning-based personalized head pose},
  author    = {Yi, Ran and Ye, Zipeng and Zhang, Juyong and Bao, Hujun and Liu, Yong-Jin},
  journal   = {arXiv:2002.10137},
  year      = {2020}
}

@article{fomm,
  title={First order motion model for image animation},
  author={Siarohin, Aliaksandr and Lathuili{\`e}re, St{\'e}phane and Tulyakov, Sergey and Ricci, Elisa and Sebe, Nicu},
  journal={NeurIPS},
  year={2019}
}

@inproceedings{fsgan,
  title={Fsgan: Subject agnostic face swapping and reenactment},
  author={Nirkin, Yuval and Keller, Yosi and Hassner, Tal},
  booktitle={ICCV},
  year={2019}
}

@InProceedings{ffpp,
author = {R\"ossler, Andreas and Cozzolino, Davide and Verdoliva, Luisa and Riess, Christian and Thies, Justus and Niessner, Matthias},
title = {FaceForensics++: Learning to Detect Manipulated Facial Images},
booktitle = {ICCV},
year = {2019}
}

@InProceedings{cdf,
author = {Li, Yuezun and Yang, Xin and Sun, Pu and Qi, Honggang and Lyu, Siwei},
title = {Celeb-DF: A Large-Scale Challenging Dataset for DeepFake Forensics},
booktitle = {CVPR},
year = {2020}
}

@article{cdfpp,
  title={Celeb-DF++: A Large-scale Challenging Video DeepFake Benchmark for Generalizable Forensics},
  author={Li, Yuezun and Zhu, Delong and Cui, Xinjie and Lyu, Siwei},
  journal={arXiv:2507.18015},
  year={2025}
}

@article{dfdcp,
  title={The deepfake detection challenge (dfdc) preview dataset},
  author={Dolhansky, Brian and Howes, Russ and Pflaum, Ben and Baram, Nicole and Ferrer, Cristian Canton},
  journal={arXiv:1910.08854},
  year={2019}
}

@article{dfdc,
  title={The deepfake detection challenge dataset},
  author={Dolhansky, Brian and Bitton, Joanna and Pflaum, Ben and Lu, Jikuo and Howes, Russ and Wang, Menglin and Canton Ferrer, Cristian},
  journal={arXiv:2006.07397},
  year={2020}
}

@inproceedings{deeperforensics,
  title={{DeeperForensics-1.0}: A Large-Scale Dataset for Real-World Face Forgery Detection},
  author={Jiang, Liming and Li, Ren and Wu, Wayne and Qian, Chen and Loy, Chen Change},
  booktitle={CVPR},
  year={2020}
}

@inproceedings{wilddf,
  title={Wilddeepfake: A challenging real-world dataset for deepfake detection},
  author={Zi, Bojia and Chang, Minghao and Chen, Jingjing and Ma, Xingjun and Jiang, Yu-Gang},
  booktitle={MM},
  year={2020}
}

@InProceedings{ffiw,
    author    = {Zhou, Tianfei and Wang, Wenguan and Liang, Zhiyuan and Shen, Jianbing},
    title     = {Face Forensics in the Wild},
    booktitle = {CVPR},
    year      = {2021},
}

@Misc{dfd,
  howpublished = "\url{https://ai.googleblog.com/2019/09/contributing-data-to-deepfake-detection.html}",
  note={Accessed: 2025-11-4},
  key = {Contributing data to deepfake detection research}
}

@Misc{sora,
  howpublished = "\url{https://sora.chatgpt.com/}",
  note={Accessed: 2025-11-4},
  key = {Sora2}
}

@article{avspeech,
  title={Looking to listen at the cocktail party: a speaker-independent audio-visual model for speech separation},
  author={Ephrat, Ariel and Mosseri, Inbar and Lang, Oran and Dekel, Tali and Wilson, Kevin and Hassidim, Avinatan and Freeman, William T and Rubinstein, Michael},
  journal={SIGGRAPH},
  year={2018},
}

@inproceedings{voxceleb2,
  title={Voxceleb2: Deep speaker recognition},
  author={Chung, Joon Son and Nagrani, Arsha and Zisserman, Andrew},
  booktitle={INTERSPEECH},
  year={2018}
}

@INPROCEEDINGS{acappella,
  author={A cappella: Audio-visual Singing Voice Separation},
  booktitle={BMVC}, 
  title={A 3D Face Model for Pose and Illumination Invariant Face Recognition}, 
  year={2021}
}

@article{dftimit,
  author       = {Pavel Korshunov and S{\'{e}}bastien Marcel},
  title        = {DeepFakes: a New Threat to Face Recognition? Assessment and Detection},
  journal      = {arXiv:1812.08685},
  year         = {2018},
}

@TECHREPORT{Sanderson2002TheVD,
                      author = {Sanderson, Conrad},
                    projects = {Idiap},
                       title = {{T}he {V}id{TIMIT} {D}atabase},
                        type = {Idiap-Com},
                      number = {Idiap-Com-06-2002},
                        year = {2002},
                 institution = {IDIAP},
ipdmembership={learning},
}

@INPROCEEDINGS{uadfv,
    author={Yuezun Li, Ming-ching Chang and Siwei Lyu},
    title={In Ictu Oculi: Exposing AI Generated Fake Face Videos by Detecting Eye Blinking},
    booktitle={WIFS},
    year={2018}
}

@inproceedings{favc,
    author = {Khalid, Hasam and Kim, Minha and Tariq, Shahroz and Woo, Simon S.},
    title = {Evaluation of an Audio-Video Multimodal Deepfake Dataset using Unimodal and Multimodal Detectors},
    year = {2021},
    booktitle = {MM Workshop},
}

@inproceedings{avdf1m,
  title={AV-Deepfake1M: A large-scale LLM-driven audio-visual deepfake dataset},
  author={Cai, Zhixi and Ghosh, Shreya and Adatia, Aman Pankaj and Hayat, Munawar and Dhall, Abhinav and Gedeon, Tom and Stefanov, Kalin},
  booktitle={MM},
  year={2024}
}

@inproceedings{lavdf,
  title={Do you really mean that? content driven audio-visual deepfake dataset and multimodal method for temporal forgery localization},
  author={Cai, Zhixi and Stefanov, Kalin and Dhall, Abhinav and Hayat, Munawar},
  booktitle={DICTA},
  year={2022},
}

@inproceedings{kodf,
  title={Kodf: A large-scale korean deepfake detection dataset},
  author={Kwon, Patrick and You, Jaeseong and Nam, Gyuhyeon and Park, Sungwoo and Chae, Gyeongsu},
  booktitle={ICCV},
  year={2021}
}

@inproceedings{dfbench,
  title={DeepfakeBench: a comprehensive benchmark of deepfake detection},
  author={Yan, Zhiyuan and Zhang, Yong and Yuan, Xinhang and Lyu, Siwei and Wu, Baoyuan},
  booktitle={NeurIPS},
  year={2023}
}

@inproceedings{idforge,
  title={Identity-Driven Multimedia Forgery Detection via Reference Assistance},
  author={Xu, Junhao and Chen, Jingjing and Song, Xue and Han, Feng and Shan, Haijun and Jiang, Yu-Gang},
  booktitle={MM},
  year={2024}
}

@article{multiface,
  title={Multiface: A dataset for neural face rendering},
  author={Wuu, Cheng-hsin and Zheng, Ningyuan and Ardisson, Scott and Bali, Rohan and Belko, Danielle and Brockmeyer, Eric and Evans, Lucas and Godisart, Timothy and Ha, Hyowon and Huang, Xuhua and others},
  journal={arXiv:2207.11243},
  year={2022}
}

@article{flame, 
  title = {Learning a model of facial shape and expression from {4D} scans}, 
  author = {Li, Tianye and Bolkart, Timo and Black, Michael. J. and Li, Hao and Romero, Javier}, 
  journal = {SIGGRAPH Asia}, 
  year = {2017},
}

@article{spectre,
  title={Visual Speech-Aware Perceptual 3D Facial Expression Reconstruction from Videos},
  author={Filntisis, Panagiotis P. and Retsinas, George and Paraperas-Papantoniou, Foivos and Katsamanis, Athanasios and Roussos, Anastasios and Maragos, Petros},
  journal={arXiv:2207.11094},
  year={2022}
}

@inproceedings{facediffuser, 
author = {Stan, Stefan and Haque, Kazi Injamamul and Yumak, Zerrin},
title = {FaceDiffuser: Speech-Driven 3D Facial Animation  Synthesis Using Diffusion},
year = {2023},
booktitle = {SIGGRAPH}
}

@inproceedings{voca, 
    title = {Capture, Learning, and Synthesis of {3D} Speaking Styles},
    author = {Cudeiro, Daniel and Bolkart, Timo and Laidlaw, Cassidy and Ranjan, Anurag and Black, Michael},
    booktitle = {CVPR},
    year = {2019}
}

@InProceedings{imitator, 
        author    = {Thambiraja, Balamurugan and Habibie, Ikhsanul and Aliakbarian, Sadegh and Cosker, Darren and Theobalt, Christian and Thies, Justus}, 
        title     = {Imitator: Personalized Speech-driven 3D Facial Animation}, 
        booktitle = {ICCV}, 
        year = {2023} 
}

@inproceedings{codetalker,
  title={Codetalker: Speech-driven 3d facial animation with discrete motion prior},
  author={Xing, Jinbo and Xia, Menghan and Zhang, Yuechen and Cun, Xiaodong and Wang, Jue and Wong, Tien-Tsin},
  booktitle={CVPR},
  year={2023}
}

@inproceedings{emote,
  title = {Emotional Speech-Driven Animation with Content-Emotion Disentanglement},
  author = {Daněček, Radek and Chhatre, Kiran and Tripathi, Shashank and Wen, Yandong and Black, Michael and Bolkart, Timo},
  year = {2023},
  booktitle={SIGGRAPH Asia}
}

@inproceedings{gaussianavatars,
  title={Gaussianavatars: Photorealistic head avatars with rigged 3d gaussians},
  author={Qian, Shenhan and Kirschstein, Tobias and Schoneveld, Liam and Davoli, Davide and Giebenhain, Simon and Nie{\ss}ner, Matthias},
  booktitle={CVPR},
  year={2024}
}

@inproceedings{headstudio,
  title = {HeadStudio: Text to Animatable Head Avatars with 3D Gaussian Splatting},
  author = {Zhenglin Zhou and Fan Ma and Hehe Fan and Zongxin Yang and Yi Yang},
  booktitle = {ECCV},
  year={2024},
}

@article{catastrophic,
  title={Overcoming catastrophic forgetting in neural networks},
  author={Kirkpatrick, James and Pascanu, Razvan and Rabinowitz, Neil and Veness, Joel and Desjardins, Guillaume and Rusu, Andrei A and Milan, Kieran and Quan, John and Ramalho, Tiago and Grabska-Barwinska, Agnieszka and others},
  journal={PNAS},
  year={2017},
}

@INPROCEEDINGS{CanTheyBeGeneralized,
  author={Khodabakhsh, Ali and Ramachandra, Raghavendra and Raja, Kiran and Wasnik, Pankaj and Busch, Christoph},
  booktitle={BIOSIG}, 
  title={Fake Face Detection Methods: Can They Be Generalized?}, 
  year={2018}
  }

@inproceedings{xuan2019generalization,
  title={On the generalization of GAN image forensics},
  author={Xuan, Xinsheng and Peng, Bo and Wang, Wei and Dong, Jing},
  booktitle={CCBR},
  year={2019}
}

@article{cozzolino2018forensictransfer,
  title={Forensictransfer: Weakly-supervised domain adaptation for forgery detection},
  author={Cozzolino, Davide and Thies, Justus and R{\"o}ssler, Andreas and Riess, Christian and Nie{\ss}ner, Matthias and Verdoliva, Luisa},
  journal={arXiv:1812.02510},
  year={2018}
}

@inproceedings{lae,
  title={Towards generalizable deepfake detection with locality-aware autoencoder},
  author={Du, Mengnan and Pentyala, Shiva and Li, Yuening and Hu, Xia},
  booktitle={CIKM},
  year={2020}
}

@inproceedings{mish,
  title={Mish: A self regularized non-monotonic activation function},
  author={Misra, Diganta},
  booktitle={BMVC},
  year={2020}
}

@article{paszke2019pytorch,
  title={Pytorch: An imperative style, high-performance deep learning library},
  author={Paszke, Adam and Gross, Sam and Massa, Francisco and Lerer, Adam and Bradbury, James and Chanan, Gregory and Killeen, Trevor and Lin, Zeming and Gimelshein, Natalia and Antiga, Luca and others},
  journal={NeurIPS},
  year={2019}
}

@inproceedings{fan,
  title={How far are we from solving the 2D \& 3D Face Alignment problem? (and a dataset of 230,000 3D facial landmarks)},
  author={Bulat, Adrian and Tzimiropoulos, Georgios},
  booktitle={ICCV},
  year={2017}
}
}

\end{document}